\documentclass{article}

\PassOptionsToPackage{numbers, compress}{natbib}

\usepackage[main, final]{neurips_2026}

\usepackage[utf8]{inputenc} 
\usepackage[T1]{fontenc}    
\usepackage{hyperref}       
\usepackage{url}            
\usepackage{booktabs}       
\usepackage{amsfonts}       
\usepackage{nicefrac}       
\usepackage{microtype}      
\usepackage[svgnames, dvipsnames]{xcolor}         
\usepackage{graphicx}       
\usepackage{amsmath}
\usepackage{enumitem}
\usepackage{multirow}
\usepackage{subcaption} 
\usepackage{amssymb}

\usepackage{wrapfig}  
\usepackage{tikz}     

\newif\ifshowlayoutchanges \showlayoutchangestrue
\newsavebox{\layoutchgbox}
\newenvironment{layoutchg}{%
  \begin{lrbox}{\layoutchgbox}\begin{minipage}{\linewidth}}
 {\end{minipage}\end{lrbox}%
  \usebox{\layoutchgbox}%
  \ifshowlayoutchanges{\color{red}\setlength{\unitlength}{1sp}%
    \begin{picture}(0,0)
      \put(\number\numexpr-\wd\layoutchgbox-131072\relax,\number\numexpr-\dp\layoutchgbox-131072\relax){%
        \framebox(\number\numexpr\wd\layoutchgbox+262144\relax,%
                  \number\numexpr\ht\layoutchgbox+\dp\layoutchgbox+262144\relax){}}%
    \end{picture}}\fi}

\newif\ifshowlayoutchanges \showlayoutchangesfalse
\usepackage[normalem]{ulem}
\newif\ifshowtextchanges \showtextchangestrue
\newif\ifshowdeletions \showdeletionsfalse
\newcommand{\add}[1]{\ifshowtextchanges{\color{blue}#1}\else#1\fi}
\newcommand{\addr}[1]{\ifshowtextchanges{\color{red}#1}\else#1\fi}
\newcommand{\del}[1]{\ifshowdeletions{\color{red}\sout{#1}}\fi}
\newcommand{\addbegin}{\ifshowtextchanges\color{blue}\fi}
\newcommand{\addend}{\ifshowtextchanges\normalcolor\fi}

\newif\ifshowtextchanges \showtextchangesfalse

\title{Learning Visual Feature-Based World Models via Residual Latent Action}

\author{
  \textbf{Xinyu Zhang}$^{1}$ \quad
  \textbf{Zhengtong Xu}$^{2}$ \quad
  \textbf{Yutian Tao}$^{3}$ \\[1ex] 
  \textbf{Yeping Wang}$^{3}$ \quad
  \textbf{Yu She}$^{2}$ \quad
  \textbf{Abdeslam Boularias}$^{1}$ \\[2ex]
  $^1$Rutgers University \qquad $^2$Purdue University \qquad $^3$University of Wisconsin-Madison
}

\begin{document}

\maketitle

\begin{abstract}
World models predict future transitions from observations and actions. Existing works predominantly focus on image generation only. Visual feature-based world models, on the other hand, predict future visual features instead of raw video pixels, offering a promising alternative that is more efficient and less prone to hallucination.  However, current feature-based approaches rely on direct regression, which leads to blurry or collapsed predictions in complex interactions, while generative modeling in high-dimensional feature spaces still remains challenging. 
In this work, we discover that a new type of latent action representation, which we refer to as {\it Residual Latent Action} (RLA), can be easily learned from DINO residuals. We also show that RLA is predictive, generalizable, and encodes temporal progression. 
Building on RLA, we propose {\it RLA World Model} (RLA-WM), which predicts RLA values via flow matching. RLA-WM outperforms both state-of-the-art feature-based and video-diffusion world models on simulation and real-world datasets, while being orders of magnitude faster than video diffusion. 
Furthermore, we develop two robot learning techniques that use RLA-WM  to improve policy learning. The first one is a minimalist world action model with RLA that learns from actionless videos\add{, and improves  VLA on LIBERO and real robot}. The second one is \del{the first}\add{a} visual RL framework trained entirely inside a world model learned from offline videos only, using a video-aligned reward and no online interactions\del{ or handcrafted rewards}.


\end{abstract}
\section{Introduction}



\let\thefootnote\relax\footnotetext{Project page: \url{https://mlzxy.github.io/rla-wm}}

World models have recently received increasing research attention due to their  great potential for policy learning and reasoning through future state prediction~\cite{dong2026learning}. 
Currently, the predominant paradigm in world modeling relies on video generation, by predicting future trajectories in pixel-aligned VAE latent spaces~\cite{yang2024learning, wu2025rlvr, wu2025geometry, wang2025precise, huang2025vid2world, cen2025worldvla}. While visually compelling, this approach is prone to hallucination~\cite{han2025video} and suffers from a heavy computational overhead~\cite{zheng2025open}. As a result, downstream applications of world models remain largely constrained to open-loop robot data generation~\citep{wang2026interactive, team2025gigabrain}, policy pretraining~\citep{mendonca2023structured, lu2025gwm}, and planning for specific tasks~\citep{du2024video, chen2025planning, zhou2025dino}.

Visual feature-based world models predict features of future frames, such as DINO tokens, rather than just videos~\cite{assran2025v}. This direction is partly motivated by studies in cognitive science showing that humans do not reason in raw pixels but in latent spaces shaped by task goals and physical understanding~\cite{yamins2016using, battaglia2013simulation}. DINO-WM~\cite{zhou2025dino, chun2025sparse, baldassarre2025back} shows that direct regression of future DINO tokens leads to efficient and accurate world models for 2D manipulation tasks. However, despite these advantages, feature-based world models remain far less adopted, as predictions often become blurry or even collapse in complex 3D interactions~\cite{wang2026interactive}.
A seemingly straightforward solution is to use generative models in feature space. However, feature-space generation is even more difficult than in pixel space due to the higher dimensionality~\cite{zheng2025rethinking, zheng2025diffusion}. More importantly, heavy generative pipelines undermine the very advantages that feature-based models should provide, as detailed in Sec.~\ref{sec:method}.

Motivated by these challenges, we seek to answer two key questions: (1) how to develop an efficient yet accurate world model in a visual feature space that scales to complex 3D manipulation? and (2) how to leverage such world models to improve downstream policies?

\begin{figure}[t]
    \centering
   \begin{minipage}{1.0\linewidth}
    \begin{subfigure}[b]{0.32\textwidth}
        \centering
        \includegraphics[height=3.9cm, keepaspectratio]{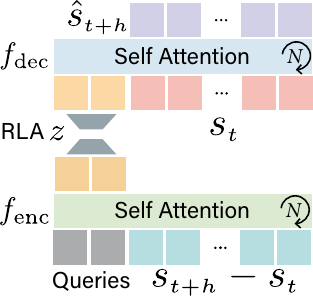}
        \caption{RLA Autoencoder Learning}
        \label{fig:rla-1}
    \end{subfigure}
    \hfill 
    \begin{subfigure}[b]{0.32\textwidth}
        \centering
        \includegraphics[height=3.9cm, keepaspectratio]{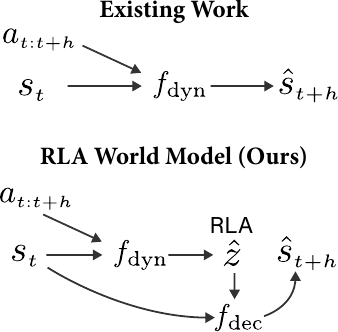}
        \caption{Dynamics Learning}
        \label{fig:rla-2}
    \end{subfigure}
    \hfill 
    \begin{subfigure}[b]{0.32\textwidth}
        \centering
        \includegraphics[height=3.9cm, keepaspectratio]{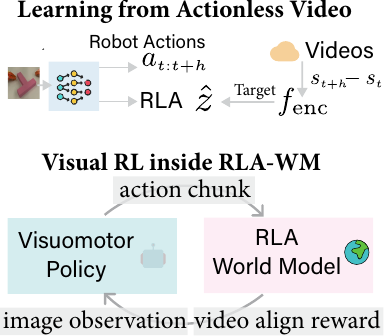}
        \caption{Applications}
        \label{fig:rla-3}
    \end{subfigure}
    \end{minipage}
    \caption{\textbf{Overview of our framework.} We introduce the Residual Latent Action (RLA), which compresses the DINO token residual $s_{t+h} - s_t$ into a compact latent $z$. We discover that RLA is predictive, generalizable, and encodes temporal progression. Next, we propose the RLA World Model (RLA-WM), which learns from offline videos and predicts RLA $z$ instead of $s_{t+h}$ directly. RLA-WM achieves accurate future prediction while being more efficient than state-of-the-art feature-based and video diffusion world models.   Our approach enables two applications: learning policies from actionless videos and visual reinforcement learning via interaction entirely within RLA-WM.}
    \label{fig:rla}
\end{figure}

While visual features are high-dimensional, we believe the manifold of valid physical transitions is inherently lower-dimensional. Therefore, learning a compact representation of these low-dimensional dynamics would enable a more principled approach to visual feature-based world models. 
In this work, we introduce {\it Residual Latent Actions} (RLA). RLA is deceptively simple: it encodes the residual between DINO tokens of two frames $(s_t, s_{t+h})$ into a compact latent vector, and is trained with a single regression loss to reconstruct $s_{t+h}$ from $s_{t}$, as shown in Fig.~\ref{fig:rla-1}.
Despite its simplicity, we find that RLA exhibits three surprising empirical properties that make it well-suited for dynamics learning.
(1) RLA is sufficiently predictive. As illustrated in Fig.~\ref{fig:rla-predictive}, the decoder $f_\text{dec}$ can accurately reconstruct $s_{t+h}$ from RLA $z$ and $s_t$ in a single forward pass. In contrast, prior methods mainly use latent actions as weak conditioning labels for iterative generation~\cite{gao2025adaworld, bi2025motus}. (2) RLA generalizes to novel scenes and motion patterns, even when trained on limited data, as shown in Fig.~\ref{fig:rla-generalizable}.
(3) RLA latent space exhibits a temporal topology; although training is performed only on frame pairs, decoding linear interpolations between a Gaussian noise and RLA yields results that approximate intermediate frames, as illustrated in Fig.~\ref{fig:rla-topology}.


Based on RLA, we propose the {\it RLA World Model} (RLA-WM), shown in Fig.~\ref{fig:wm}. Instead of directly regressing DINO tokens $s_{t+h}$, RLA-WM first predicts RLA $z$ via flow matching with $s_t$ and actions $a_{t:t+h}$ as input conditions, then predicts $s_{t+h}$ from $s_{t}$ and $z$.
RLA-WM significantly outperforms state-of-the-art feature-based and video diffusion world models on both simulation and real-world datasets, while remaining more efficient as the flow matching runs in the compact RLA space.


Furthermore, we introduce two robot learning techniques built on RLA and RLA-WM. First, we show that a behavior cloning policy can be extended into a minimalist world action model (WAM) using a single linear layer that predicts RLA from the current observation.
Unlike prior WAMs that couple action prediction with heavy video generation backbones~\cite{cen2025worldvla}, our approach imposes no such coupling, adds no inference cost, and consistently improves policy success rates for imitation learning from actionless videos\add{, with further gains for a VLA on LIBERO~\cite{liu2023libero} and a real robot cup stacking task}.
Second, we \del{present the first demonstration of}\add{perform} visual reinforcement learning (RL) entirely inside a world model learned from a small offline video dataset without online interactions\del{,  handcrafted rewards,} or even auxiliary BC loss during RL\add{, using an alignment reward with offline videos}.
Our World Model-based RL (WMRL)  \del{yields a significant improvement on ManiSkill tasks for the XArm and UR10e robots}\add{improves the best BC policies on all five ManiSkill tasks, with statistically significant gains on the XArm and UR10e robots, and further benefits from failed trajectories that BC cannot exploit}.

Our contributions are threefold.
    (1) We propose the Residual Latent Action (RLA), a simple latent action representation learned from DINO residuals. 
    (2) We present RLA-WM, which predicts RLA via flow matching and sets a new state-of-the-art among visual feature-based world models.
    (3) We demonstrate the value of RLA and RLA-WM in two novel applications: ($a$) a minimalist world action model that learns from actionless videos\addr{, which also improves a VLA on LIBERO and a real-robot policy};
    ($b$) a visual reinforcement learning framework that optimizes policy via rollouts in the RLA-WM\addr{, which further improves policies from failed trajectories}.

\section{Related Work}

\noindent\textbf{World Models for Robotics.} Learning world models from offline datasets has emerged as a promising paradigm for future state prediction in robotics~\citep{zhen20243d, yang2024learning}. Existing approaches largely focus on predicting future videos~\citep{wu2025rlvr, wu2025geometry, wang2025precise, huang2025vid2world, cen2025worldvla} and 3D geometry, such as point clouds~\citep{yu2025manigaussian++, huang2026pointworld, huang2025particleformer, chai2025gaf}. 
Despite their success, video prediction induces a heavy computational overhead due to diffusion models. While 3D world models benefit from spatial priors, their structural assumptions often limit them to specific tasks. Another line of research explores learning world models via online rollouts within simulators~\citep{li2025robotic, jyothir2023gradient, hansen2024td, hafner2023mastering, hafner2025training, georgiev2025pwm}, but the reliance on simulators and handcrafted reward functions limits their application. 

\noindent\textbf{World Models in Visual Feature Space.} An alternative to pixel-space prediction is embedding future states in a learned visual feature space. For instance, V-JEPA predicts future features for self-supervised learning~\citep{assran2025v, mur2026v}. The DINO-WM family of world models~\citep{zhou2025dino, chun2025sparse, baldassarre2025back} predicts DINO tokens~\citep{oquab2023dinov2, simeoni2025dinov3} of future frames through a direct regression. DINO-WM~\cite{zhou2025dino} shows that predicting in a feature space mitigates the need for heavy generative models for 2D robot manipulation tasks. However, for complex 3D manipulation, we observe that simply applying regression in the feature space often yields blurred or collapsed estimations. 
In contrast, our approach avoids regression-to-the-mean, enabling efficient and accurate multi-modal prediction of DINO tokens in future frames.

\noindent\textbf{Latent Actions.} Learning compact latent actions from videos has emerged as a popular technique in robot learning~\citep{zhang2025latent, zheng2025flare, ye2025latent, tharwat2025latent, bu2025univla, gao2025adaworld, bi2025motus, kerssies2026frame}. Existing approaches fall into two categories. The first leverages latent actions as proxy controls for imitation learning from actionless videos where proprioceptive data are absent~\citep{zheng2025flare, ye2025latent, tharwat2025latent, bu2025univla}. The second utilizes latent actions as weak condition labels for video diffusion~\cite{gao2025adaworld, bi2025motus}. In contrast, we learn Residual Latent Action (RLA) from DINO residuals instead of raw pixels. 
RLA outperforms existing methods~\cite{gao2025adaworld, bu2025univla} as a better action proxy (Sec.~\ref{sec:learn-from-actionless}), without requiring diffusion, and can be decoded into DINO tokens of future frames in a single feedforward pass.
\add{Different from LAPA~\cite{ye2025latent}, which uses latent actions to pretrain policies, RLA is designed for world modeling, i.e., predicting future states, and can be retargeted to improve policies.}





\section{Method}
\label{sec:method}

{\bf Problem Formulation.}
Let $x_t \in \mathbb{R}^{H \times W \times 3}$ denote the raw image observation at time $t$. We represent the DINO patch tokens as $s_t \in \mathbb{R}^{L \times C}$, where $C$ is the feature dimension and $L = \frac{H \times W}{P^2}$ is the sequence length for a given patch size $P$. 
We define an action chunk of horizon $h$ at time $t$  as $a_{t:t+h}$. Our objective is to learn a dynamics function $f_{\text{dyn}} : (s_t, a_{t:t+h}) \mapsto s_{t+h}$ using only raw offline videos, without online rollouts
or access to handcrafted reward functions or labels. This function acts as a direct, multi-step world model in the feature space.

{\bf Learning Latent Actions on DINO Residuals.}
The physical world is inherently uncertain, which makes the dynamics function $f_\text{dyn}$ highly multi-modal. That is, given an image $x_t$ and actions $a_{t:t+h}$, there can be multiple valid values for $x_{t+h}$. 
Prior work addresses this through generative video models to predict $\hat{x}_{t+h}$. However, these methods are computationally heavy and prone to hallucination~\cite{han2025video}.
Pioneering works such as DINO-WM and JEPA instead learn world models in a feature space, such as DINO tokens, which is more efficient, does not require diffusion, and reduces hallucination because visual features encode rich semantic and geometric information~\cite{zhou2025dino}. 
These works motivate us to design a world model that, given $s_t$, directly predicts $\hat{s}_{t+h}$  in DINO token space rather than predicting pixel-level $\hat{x}_{t+h}$.
However, despite the impressive results of DINO-WM, the key limitation is its direct regression design, which is computationally efficient but often results in blurry or collapsed predictions in complex 3D interactions.

A straightforward solution is to revert to generative models, such as diffusion or flow matching, but in the feature space, to predict $\hat{s}_{t+h}$. 
However, a counter-intuitive yet critical fact is that DINO tokens (and ViT or ResNet features generally) have a far higher dimensionality than the pixel-aligned VAE latents used in image or video generation. For a $512 \times 512$ image, Stable Diffusion VAE~\cite{esser2024scaling} yields roughly $64^2 \times 4 \approx 16$k dimensions, whereas DINOv3-L tokens produce $32^2 \times 1024 \approx 1$M dimensions, nearly two orders of magnitude larger. This curse of dimensionality makes generative modeling of DINO tokens highly challenging~ \citep{zheng2025rethinking}. While RAE~\citep{zheng2025diffusion} proposes diffusion techniques to generate DINO tokens from noise and class labels, it is not widely adopted because adapting it to a dynamics learning setting is not trivial, as shown in Tab.~\ref{tab:wm}. More importantly, using heavy generative models defeats the purpose of feature-space learning, as they undermine both the computational efficiency and the reduced hallucination that feature-based world models offer.

To address this challenge, we shift focus from directly generating $\hat{s}_{t+h}$ to learning a representation that captures the transition from $s_t$ to $s_{t+h}$.  We propose learning this representation from DINO token residuals $s_{t+h} - s_t$, which also corresponds to the flow matching velocity of a Schr\"odinger bridge~\cite{shi2023diffusion, de2024schrodinger} from $s_t$ to $s_{t+h}$. 
Specifically, we feed these residuals along with learnable queries into an encoder $f_{\text{enc}}$, project the output queries to a low-dimensional space to obtain $z$, and pass $z$ along with $s_t$ into a decoder $f_{\text{dec}}$ to reconstruct $s_{t+h}$ (Fig.~\ref{fig:rla-1}). 
We refer to $z$ as a \textit{Residual Latent Action} (RLA). The RLA autoencoder uses almost only self-attention and a single regression loss on $s_{t+h}$.
There are three key properties of RLA that set it apart from prior work, making it an ideal representation for dynamics modeling: 
\begin{description}[nosep, leftmargin=1em, font=\bfseries]
    \item[Predictive Sufficiency] Unlike in prior work, where latent actions serve only as weak conditioning for diffusion, RLA does not require iterative generation. We find that our RLA decoder $f_\text{dec}$, when conditioned on a compact RLA $z$, is able to reconstruct future DINO tokens with high fidelity in a single feedforward pass. Reconstruction examples are provided in  Fig.~\ref{fig:rla-predictive}.
    
    \item[Generalizability] RLA autoencoder generalizes to novel scenes. In Sec.~\ref{sec:learn-from-actionless}, we demonstrate this by training RLA on task-agnostic videos  and applying it to task-relevant, actionless videos for imitation learning.
    Examples of encoding unseen robot object interactions are provided in Fig.~\ref{fig:rla-generalizable}.

    \item[Temporal Topology] An emergent property of RLA is the topology of its learned latent space. Although the autoencoder is only trained on frame pairs $(s_t, s_{t+h})$, the RLA latent space naturally encodes temporal progression. Interpolating between a Gaussian noise and RLA produces frames that correspond  to temporally intermediate states, as shown in Fig.~\ref{fig:rla-topology} \add{ and quantified in Tab.~\ref{tab:interp}}.


\end{description}

{\bf RLA World Model.}
\begin{figure}[t]
    \centering
    \includegraphics[width=\linewidth]{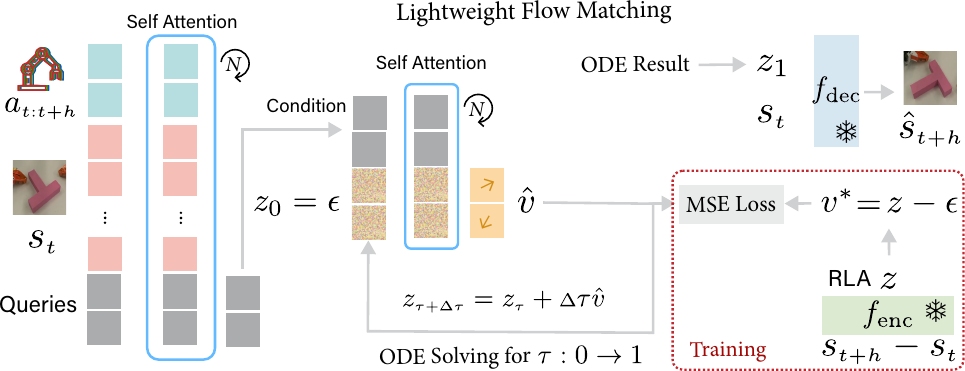}
    \caption{\textbf{RLA World Model.}  RLA-WM predicts future states by generating the residual latent action $z$. We first embed the robot actions $a_{t:t+h}$ (padded to a maximum horizon) via an MLP. This embedding is concatenated with the DINO tokens $s_t$ and learnable queries, then processed through self-attention layers to produce condition tokens. During flow matching, this condition is fixed and concatenated with a noisy latent $z_\tau$, starting from Gaussian noise $z_0 = \epsilon$. The flow network predicts the velocity $\hat{v}$ to iteratively transform $z_0$ into the final RLA $z_1$. Finally, $f_\text{dec}$ decodes $\hat{s}_{t+h}$ from $z_1$ and $s_t$. During training, the model is supervised by the MSE loss against the ground truth velocity $v^* = z - \epsilon$, requiring no feature or image reconstruction losses.}
    \label{fig:wm}
\end{figure}
Based on RLA, we revisit feature-based world modeling. Learning neural dynamics in RLA space encourages the model to capture state evolution rather than absolute states. This aligns with classical physics simulation, which models relative mesh displacements~\cite{chen2022crom}. 
Motivated by this, instead of generating high-dimensional $s_{t+h}$, we propose a world model to predict the compact RLA $z$, which is then decoded with current state $s_t$ to reconstruct $s_{t+h}$. Specifically, learnable queries are concatenated with $s_t$ and embedded actions $a_{t:t+h}$ and transformed via self-attention. These query tokens are then concatenated with a noisy RLA $z_\tau = \tau z + (1-\tau)\epsilon$, where $\epsilon\sim\mathcal{N}(0,I)$, through subsequent self-attention layers to predict the velocity $\hat{v}$. During training, we supervise with ground truth velocity $v^* = z - \epsilon$, where $z = f_{\text{enc}}(s_{t+h} - s_t)$. At inference, we sample $z_0 = \epsilon$ and solve the ODE with $z_{\tau + \Delta \tau} = z_{ \tau} + {\Delta} \tau \hat{v}$ from $\tau=0$ to $1$. The final $z_1$ is the predicted RLA, and is decoded via $f_{\text{dec}}$ with $s_t$. Since the condition network is executed once and iterative generation remains within the compact RLA space, the flow matching is lightweight, as shown by the floating point operations (FLOPs) reported in Tab.~\ref{tab:wm}. The compactness of the RLA space also helps the model to predict long-term dynamics more accurately, without over-attention to the excessive details in dense observation spaces.
Our RLA-WM framework is illustrated in Fig.~\ref{fig:wm}.

\section{Experiments}
\label{sec:experiment}

\begin{table}[t]
    \centering
    \caption{\textbf{Evaluation of Future Frame Prediction.} We evaluate RLA-WM by predicting $\hat{s}_{t+h}$ given $s_t$ and actions $a_{t:t+h}$. We report LPIPS, SSIM, L1 distance to ground-truth DINO tokens $s_{t+h}$ (DINO L1), and FLOPs per inference. Best results are \textbf{bold}, and second-best results are \underline{underlined}.}
    \label{tab:wm}
    \resizebox{\textwidth}{!}{
    \begin{tabular}{lccccccc}
        \toprule
        \multirow{2}{*}{Model} & \multicolumn{3}{c}{ManiSkill~\cite{mu2maniskill}} & \multicolumn{3}{c}{IWS~\cite{wang2026interactive}} & \multirow{2}{*}{FLOPs $\downarrow$} \\
        \cmidrule(lr){2-4} \cmidrule(lr){5-7}
        & LPIPS $\downarrow$ & SSIM $\uparrow$ & DINO L1 $\downarrow$ & LPIPS $\downarrow$ & SSIM $\uparrow$ & DINO L1 $\downarrow$ & \\
        \midrule
        DINO-WM~\cite{zhou2025dino} & 0.156 & 0.865 & 0.078 & \underline{0.223} & \underline{0.825} & \underline{0.058} & \textbf{2.1T} \\
        RAE~\cite{zheng2025diffusion} & 0.324 & 0.717 & 0.143 & 0.550 & 0.625 & 0.159 & 14.3T \\
        FM-WM~\cite{lipmanflow} & \underline{0.127} & \underline{0.890} & \underline{0.063} & 0.360 & 0.741 & 0.119 & 14.3T \\
        Vid2World~\cite{huang2025vid2world} & 0.199 & 0.705 & 0.084 & 0.388 & 0.710 & 0.139 & 1.1P \\
        \midrule
        RLA-WM (Ours) & \textbf{0.071} & \textbf{0.931} & \textbf{0.030} & \textbf{0.196} & \textbf{0.847} & \textbf{0.053} & \underline{3.5T} \\
        \bottomrule
    \end{tabular}
    }
\end{table}

Our experiments aim to answer two key questions: (1) Can RLA-WM perform accurate multi-step prediction in a visual feature space? (2) How can RLA and RLA-WM improve robot policies? To address the first question, we evaluate the RLA-WM on simulation and real-world robot manipulation videos, using image and feature prediction metrics across multi-step rollouts.
For the second one, we provide two applications of RLA and RLA-WM: (a) extending behavior cloning (BC) policies to World Action Models (WAM) via RLA, and (b) performing visual RL entirely inside an RLA-WM. 

\subsection{Prediction Quality Evaluation}
\label{exp:prediction-quality-eval}
\noindent\textbf{Datasets.} The experiments are performed on the ManiSkill simulation suite~\citep{mu2maniskill} and the IWS real-world dataset~\cite{wang2026interactive}. In ManiSkill, we adopt three robot arms (Panda with parallel gripper, XArm with Robotiq gripper, UR10 with cylinder end-effector) across five built-in tasks: Pull Cube, Pull Cube with Tool, Roll Ball, Push T, and Poke Cube. We additionally curate a task-agnostic play environment where the robot freely interacts with primitive shapes without task-specific goals. Fig.~\ref{fig:env} shows an overview of these environments. Unlike Dreamer~\cite{hafner2023mastering}, we do not use online interactions or rewards for training. We collect 1,000 successful and 500 failed episodes per ManiSkill task using pretrained state-based PPO agents, and 3,000 play videos per robot via scripted task and motion planning. For IWS, we select the three most challenging tasks (Push T, Rope Manipulation, and Open Box) using bimanual ALOHA robots, each providing over 600 human teleoperation demonstrations.

\noindent\textbf{Training and Evaluation.} 
RLA autoencoder is trained per dataset (ManiSkill and IWS), using a single model for multiple tasks and robots. Because each robot in ManiSkill and each scene in IWS has a different action space, as in Sec.~\ref{apn:details}, we train the dynamics part of RLA-WM per robot on ManiSkill using task-relevant videos, and per task on IWS. For validation on ManiSkill, we use 10 success and 10 failure episodes, unseen during training, per task. For IWS, we use the official validation set. 
During training, we randomly sample a pair $(s_t, s_{t+h})$ separated by a variable horizon $h \in [1, 15]$. The network predicts $\hat{s}_{t+h}$ from $s_t$ and actions $a_{t:t+h}$. 
All videos have a dimension of $512\times512$.
During evaluation, we condition on an initial frame and autoregressively unroll predictions for 30 steps on ManiSkill (action chunk size 10) and 60 steps on IWS (chunk size 15), requiring 3 and 4 autoregressive steps, respectively. We measure the final frame's fidelity against the ground truth using LPIPS~\cite{zhang2018lpips}, SSIM~\cite{wang2004image}, and the L1 distance of DINO tokens. 

\begin{figure}[t]
    \centering
    \makebox[\linewidth][l]{%
        \hspace*{-1.5em}
        \includegraphics[width=\linewidth]{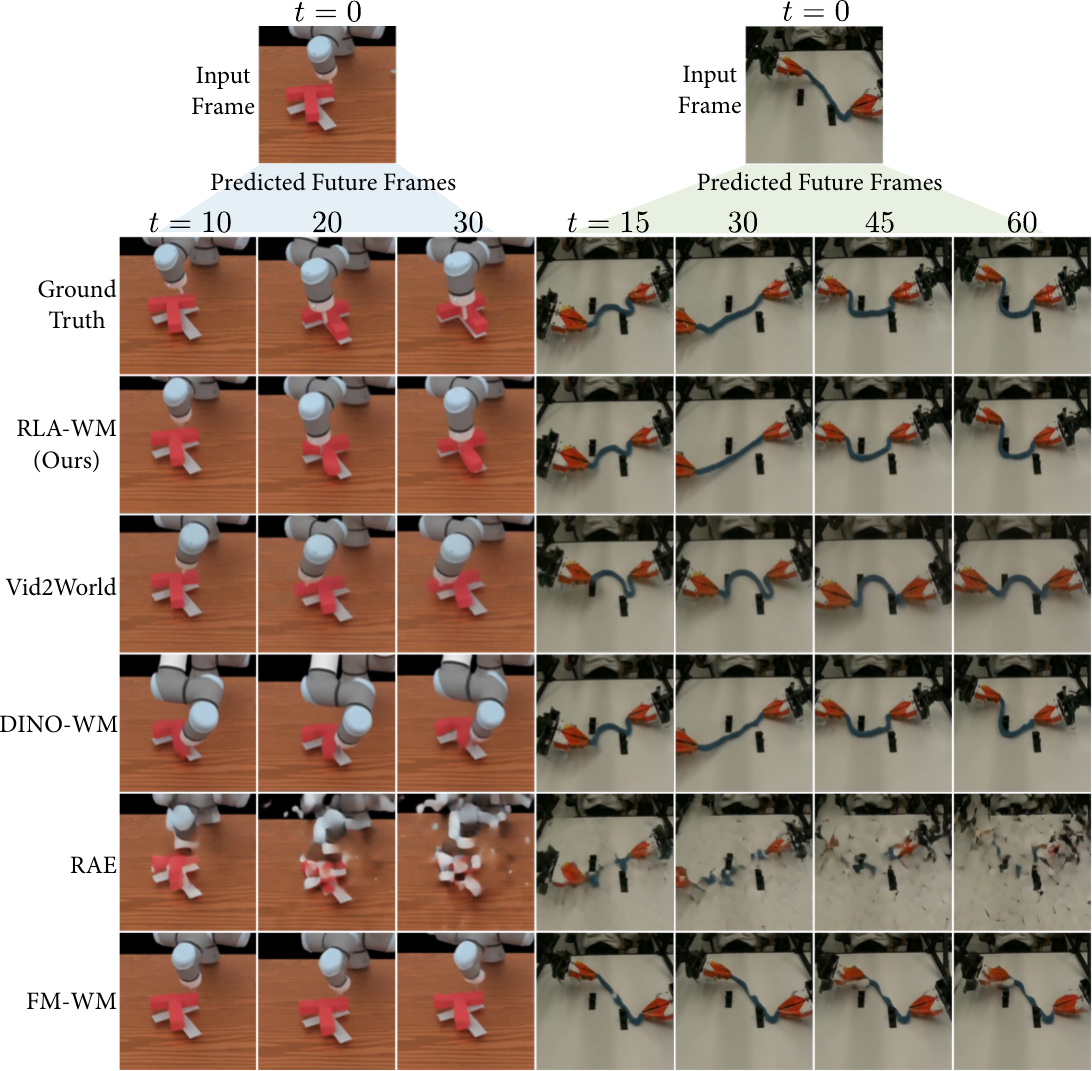}
    }
    \caption{\textbf{Qualitative Comparison for RLA-WM.} Given an input frame at $t=0$, our RLA-WM predicts future frames with high visual quality and physical fidelity, closely matching ground-truth states. In contrast, DINO-WM produces increasingly blurrier predictions for Push-T over longer horizons, and inconsistent rope states. Applying diffusion or flow matching directly in DINO token space yields inferior results (RAE, FM-WM). Vid2World generates visually sharp frames. However, it hallucinates and predicts physical states that diverge significantly from reality. }
    \label{fig:wm-result-main-text}
\end{figure}

\noindent\textbf{Baselines.} We benchmark RLA-WM against a suite of state-of-the-art visual feature-based and video diffusion-based world models: (1) \textbf{DINO-WM} \cite{zhou2025dino}, a regression network that predicts DINO tokens. We re-implement this method using DINOv3 features to regress $s_{t+h}$ given $s_t$ and an action chunk $a_{t:t+h}$; (2) \textbf{Vid2World}~\cite{huang2025vid2world}, a high-fidelity video diffusion world model based on an action-conditioned DynamicCrafter \cite{xing2024dynamicrafter} architecture with 1.1B trainable parameters; (3) \textbf{RAE} \cite{zheng2025diffusion}, a diffusion-based model for DINO tokens which we adapt to incorporate $s_t$ and $a_{t:t+h}$ as conditional inputs within our transformer backbone; (4)  \textbf{FM-WM}, a Flow Matching~\cite{lipmanflow} baseline we implement to learn a conditional probability path that directly flows $s_t$ to the future state $s_{t+h}$ given $a_{t:t+h}$.

\noindent\textbf{Results.} As detailed in Tab.~\ref{tab:wm}, RLA-WM significantly outperforms all feature-based (DINO-WM, RAE, FM-WM) and video diffusion (Vid2World) baselines across all measured metrics on both ManiSkill and IWS. We also provide qualitative comparisons on validation episodes (unseen during training) in Fig.~\ref{fig:wm-result-main-text} and Fig.~\ref{fig:wm-result-1} to \ref{fig:wm-result-5}. While Vid2World generates frames with sharp geometric and textural details, it hallucinates and predicts trajectories that lack physical grounding and diverge from reality, resulting in inferior metrics.
Furthermore, Vid2World requires 1.1P FLOPs, a computational footprint nearly three orders of magnitude larger than our 3.5T FLOPs. RLA-WM achieves high-fidelity predictions with minimal hallucination and a computational efficiency second only to the direct regression of DINO-WM. This higher performance is enabled by  RLA's ability to perform flow matching within a compact latent space.
Note that to compute image-space metrics, the DINO tokens $\hat{s}_{t+h}$  are decoded to RGB via a pre-trained UNet~\cite{ronneberger2015u}\add{, which is frozen and shared by all feature-based models. Therefore, the metric gaps stem entirely from the token prediction ability}.

\subsection{Minimalist World Action Model with RLA}
\label{sec:learn-from-actionless}

\begin{figure}[t]
    \centering
    \includegraphics[width=\linewidth]{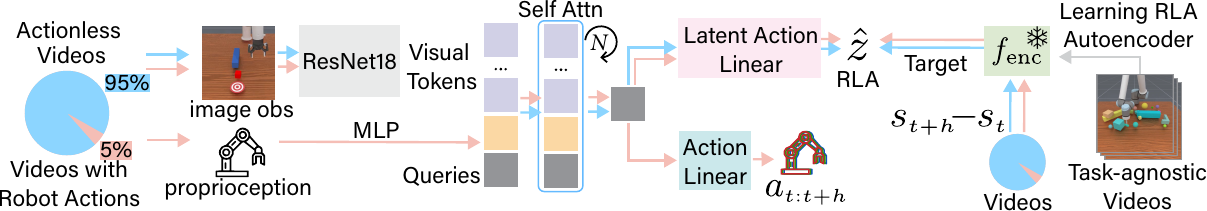}
    \caption{\textbf{Learning from Actionless Videos using RLA.} We extend a BC ResNet with a linear layer to predict the RLA $\hat{z}$. The RLA targets are extracted from $(s_t, s_{t+h})$ using an RLA encoder $f_\text{enc}$ learned from task-agnostic videos. This turns the BC policy into a minimalist world action model that can learn from videos whose proprioceptive states and robot actions are not available, without forcing the policy to couple with DINO or video generation backbones. As shown in Tab.~\ref{tab:wam}, RLA outperforms other latent action learners (replacing $f_\text{enc}$) within the same framework.}

    \label{fig:imitation}
\end{figure}

\begin{table}
  \centering
  \caption{\textbf{Latent Action Evaluation.} We report success rates (\%) for imitation policies trained on actionless videos using  latent actions. RLA achieves the highest average success rate and rank. Success rates  are evaluated over 50 episodes (seeds 42-91) and averaged over the last five checkpoints.}
  \label{tab:wam}
  \resizebox{\linewidth}{!}{
  \begin{tabular}{lccccccc}
    \toprule
    Method & PushT & Roll & Pull & Pull Tool & Poke & Rank $\downarrow$ & Avg SR $\uparrow$ \\
    \midrule
    BC-ResNet                       & 3.6  & 42.0 & 33.6 & 7.6  & 49.2 & 3.8 & 27.2 \\
    DINO CLS~\cite{simeoni2025dinov3} & 7.6  & 39.6 & 40.4 & 4.4  & 44.8 & 4.0 & 27.4 \\
    UniVLA~\cite{bu2025univla}      & 6.0  & 37.6 & 42.8 & 7.2  & 50.0 & 3.8 & 28.7 \\
    AdaWorld~\cite{gao2025adaworld} & 9.2  & 38.4 & \textbf{48.4} & 10.8 & 61.6 & 2.2 & 33.7 \\
    \midrule
    RLA (Ours)                      & \textbf{15.2} & \textbf{43.8} & 43.6 & \textbf{12.0} & \textbf{63.6} & \textbf{1.2} & \textbf{35.6} \\
    \bottomrule
  \end{tabular}}
\end{table}


\textbf{Architecture and Motivation.}
World Action Models (WAMs) combine robot
action prediction with future video generation in a hybrid architecture~\cite{cen2025worldvla, yuan2026fastwam}.
WAMs can be used as robot policies and show improved performance compared to policies trained with action prediction alone.
However, due to the complexity of video generation, existing architectures are often tightly coupled to heavy video backbones, while predicting actions via an auxiliary  module.
This coupling limits their flexibility.
The proposed RLA model provides a flexible alternative. 
We propose a minimalist WAM by extending a standard ResNet-18~\cite{he2016deep} behavior cloning (BC) policy. We first pre-train the RLA autoencoder entirely on task-agnostic play data. The BC network then takes a $128\times 128$ image observation and proprioceptive joint angles as input, projecting them into a shared feature. 
Next, the network branches into two linear heads, one predicts robot actions, and the other predicts the RLA $z$, which is supervised by the pre-trained RLA autoencoder.
Our architecture is visualized in Fig.~\ref{fig:imitation}.

\textbf{Learning from Actionless Video.} We evaluate our minimalist WAM in a setting where only a small fraction of demonstrations contain action labels -- a practical setup for scaling robot learning to large-scale, unlabeled videos. We only include robot actions and proprioceptive states for 5\% of all videos (15\% for PushT due to its difficulty). The remainder are actionless, video-only trajectories.
During training, we construct each batch by sampling equally from videos with and without actions.
 For actionless videos, we mask the action loss, replace the proprioceptive input with a learnable default token, and train the shared backbone using the RLA $z$ encoded from $(s_t, s_{t+h})$. During evaluation, we discard the RLA head and evaluate the policy's success rate using only the action head.


\textbf{Baselines.} We benchmark RLA against several latent action formulations: (1) \textbf{DINO CLS}, which uses the DINO class token of the future frame $s_{t+h}$ as the latent action. This follows Flare~\cite{zheng2025flare}, which aligns visual features $s_t$ to $s_{t+h}$. Note that Flare's implementation is not publicly accessible. (2) \textbf{UniVLA}~\cite{bu2025univla}, which learns latent actions from DINO tokens of frame pairs $(s_t, s_{t+h})$ using  spatial-temporal attention and VQ-VAE~\cite{van2017neural}.  (3) \textbf{AdaWorld}~\cite{gao2025adaworld}, which derives latent actions from frame pairs $(x_t, x_{t+h})$ via a VAE and operates on raw RGB images with cross-attention.
For evaluation, we report success rates over 50 evaluation episodes with standard seeds from 42 to 91.
Each baseline's latent extractor is pre-trained on the same dataset and substituted as the  $f_\text{enc}$ within our WAM framework in Fig.~\ref{fig:imitation}.

\textbf{Results.} Table~\ref{tab:wam} presents quantitative comparisons. Our RLA as a latent action significantly outperforms existing methods, achieving +8.5\% over the BC baseline and +1.9\% over AdaWorld (second best), yielding the highest success rate on nearly all tasks. Notably, on the most challenging Push-T task, which requires spatial reasoning and long horizons, our method shows the largest improvement (15.2\% over the baseline's 3.6\%), while  AdaWorld (second best) achieves only 9.2\%. 
\add{Since all latent actions share the same auxiliary head, the gain comes from the RLA target itself instead of architecture.}
The unique aspect of RLA is that adding one linear layer to predict $z$ makes the simple framework (Fig.~\ref{fig:imitation}) a \del{true }world action model, as RLA enables accurate prediction of $s_{t+h}$ from $s_t$. 

\add{\textbf{Scaling to VLAs and Real Robots.} We further evaluate RLA with VLA-Adapter~\cite{wang2026vla} on the LIBERO benchmark~\cite{liu2023libero}, and compare with the latent-action VLAs LAPA~\cite{ye2025latent} and UniVLA~\cite{bu2025univla}. Since LIBERO includes a moving wrist camera, we extend RLA autoencoder to concatenated DINO tokens of the front and wrist cameras as in Fig.~\ref{fig:rla-vla-real}. VLA-Adapter predicts RLA chunks alongside robot actions. As in Tab.~\ref{tab:libero}, RLA improves VLA-Adapter on Spatial and Object tasks, and achieves the highest average success rate. On a real SO-101 robot, we train BC policies for cup stacking with 120 teleoperated demonstrations from the front and wrist cameras. Adding RLA raises the success rate from 15\% to 40\% over 20 rollouts, as in Tab.~\ref{tab:real}. These results show that RLA remains effective with a moving wrist camera, transfers to VLA architecture and real-world robot learning.}




\begin{table}[t]
\begin{layoutchg}
  \centering
  \begin{minipage}[t]{0.565\linewidth}
    \centering
    \caption{\textbf{Latent-Action VLAs on LIBERO.} We report success rates (\%). We extend VLA-Adapter to also predict RLA chunks of front and wrist cameras. The policy architecture is illustrated in Fig.~\ref{fig:rla-vla-real}.}
    \label{tab:libero}
    \footnotesize
    \setlength{\tabcolsep}{3pt}
    \begin{tabular}{lccccc}
      \toprule
      Method & Spatial & Object & Goal & Long & Avg \\
      \midrule
      LAPA~\cite{ye2025latent}          & 73.8 & 74.6 & 58.8 & 55.4 & 65.7 \\
      UniVLA~\cite{bu2025univla}        & 96.5 & 96.8 & 95.6 & 92.0 & 95.2 \\
      VLA-Adapter~\cite{wang2026vla}    & 97.0 & 97.4 & \textbf{96.6} & \textbf{93.4} & 96.1 \\
      \midrule
      VLA-Adapter + RLA (Ours)          & \textbf{97.6} & \textbf{99.2} & \textbf{96.6} & \textbf{93.4} & \textbf{96.7} \\
      \bottomrule
    \end{tabular}
  \end{minipage}\hfill
  \begin{minipage}[t]{0.415\linewidth}
    \centering
    \caption{\textbf{Real-Robot Cup Stacking.} We train policies with 120 demonstrations on an \mbox{SO-101} robot (Fig.~\ref{fig:rla-vla-real}), and report success over 20 rollouts.}
    \label{tab:real}
    \footnotesize
    \setlength{\tabcolsep}{3.5pt}
    \raisebox{-0.5\height}{\includegraphics[height=1.55cm, trim=99 13.5 17.5 0, clip]{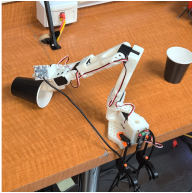}}\hspace{3pt}%
    \begin{tabular}[c]{lc}
      \toprule
      Method & Success \\
      \midrule
      BC              & 3/20 (15\%) \\
      BC + RLA (Ours) & \textbf{8/20 (40\%)} \\
      \bottomrule
    \end{tabular}
  \end{minipage}
\end{layoutchg}
\end{table}

\begin{figure}[t]
    \centering
    \includegraphics[width=\linewidth]{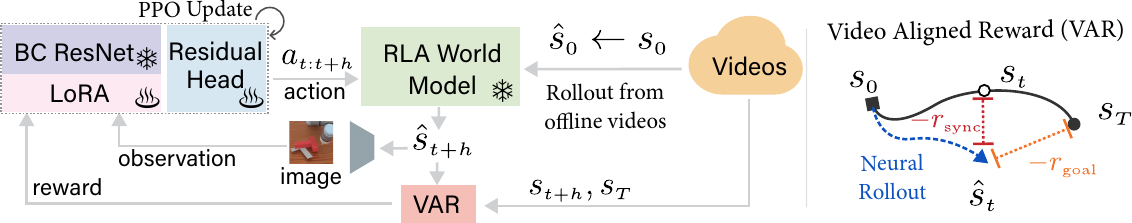}
    \caption{\textbf{Visual Reinforcement Learning within RLA World Models.}  We adapt a pretrained ResNet BC policy for RL using LoRA adapters and a residual action head predicting delta actions and a Gaussian log deviation. The policy outputs action chunks $a_{t:t+h}$ to our RLA-WM, which predicts future tokens $\hat{s}_{t+h}$. A pretrained UNet decodes $\hat{s}_{t+h}$ into RGB observations for next step. RLA-WM resets its state $\hat{s}_0$ from the initial frame $s_0$ of a randomly sampled offline demonstration. For reward, we propose Video Aligned Reward (VAR). Because neural rollouts are time-synchronized with the reference video, VAR is simply defined as the negative DINO L1 distance between $\hat{s}_t$ and $s_t$ (or terminal $s_T$). Policy is optimized via PPO using rollouts generated entirely inside RLA-WM.}
    \label{fig:wmrl}
\end{figure}

\subsection{Visual Reinforcement Learning within RLA World Model}
\label{exp:visual-rl}

\textbf{Motivation.}  Extracting a robust visuomotor policy from a world model trained on offline videos remains a fundamental challenge~\cite{ha2018world}. Existing methods generally fall into two paradigms: reinforcement learning (RL) and planning. In RL, GWM~\cite{lu2025gwm} improves sample efficiency but still requires simulator interaction.  UniSim~\cite{yang2024learning} uses a video diffusion model as a simulator, but requires massive compute, remains unreleased, and is evaluated on a single task. Conversely, planning with world models lacks a unified framework: gradient-based methods suffer from non-convex optimization landscapes~\cite{zhou2025dino}, and heuristic planners~\cite{berg2025semantic, chen2025planning} are highly task-specific.  While principled methods like TD-MPC~\cite{hansen2024td, hansen2022temporal} and the Dreamer~\cite{hafner2023mastering} exist, they rely on pre-defined reward functions and large-scale online interactions. Thus, performing RL entirely within a learned world model is a critical open problem. This requires a world model that (1) accurately predicts the future, (2) supports efficient rollouts, and (3) operates without manual reward engineering.

\textbf{Architecture.} We introduce World Model-based RL (WMRL), a framework for visual RL fully within our RLA-WM. The RL policy is initialized from a pre-trained BC-ResNet policy, with LoRA adapters and a residual head added to predict Gaussian action distributions. The policy maps an RGB image at time $t$ to an action chunk $a_{t:t+h}$, which the RLA-WM uses to predict $\hat{s}_{t+h}$. Previous works often employ SAC~\cite{haarnoja2018soft, lireinforcement} or spline~\cite{litop} to ensure a strict multi-step RL formulation. We simply adopt PPO~\cite{schulman2017proximal}, treat the $(s_t, a_{t:t+h}, s_{t+h})$ tuple as a single transition, and bypass intermediate advantage estimation. We propose to rollout from offline videos by setting $\hat{s}_0$ to the first frame $s_0$ of a 
sampled demonstration. For reward, we introduce Video Aligned Reward. Since each rollout is time-aligned with a reference offline video, we compute the reward as the negative L1 distance between the DINO tokens $\hat{s}_t$ and the ground-truth $s_t$ (or the terminal  $s_T$). We observe a neural-to-sim gap between images decoded by  UNet and the simulator's ray tracing. We reduce this gap by applying the UNet decoding as a preprocessing step.  Our WMRL framework is outlined in Fig.~\ref{fig:wmrl}.

\begin{figure}[t]
\begin{layoutchg}
    \centering
    \includegraphics[width=\linewidth]{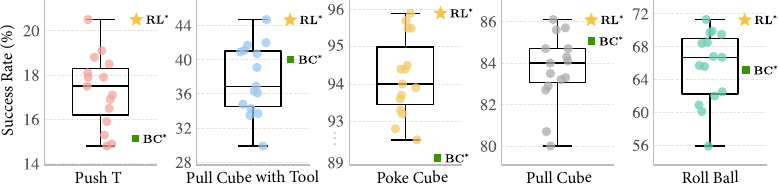}
    \caption{\textbf{World Model RL Performance Distribution.} We select the best-performing BC models ($\textcolor{DarkGreen}{\blacksquare} \ \text{BC}^*$) and apply WMRL (Fig.~\ref{fig:wmrl}) across 15 independent seeds (1-15). Each seed's best checkpoint success rate is plotted as a dot ($\textcolor{gray}{\Huge \bullet}$), with the overall optimal checkpoint marked as $\textcolor{Dandelion}{\bigstar} \ \text{RL}^*$. Success rates are evaluated on 750 episodes (seeds 1-750), according to  Tab.~\ref{tab:wmrl}.}
    \label{fig:wmrl-boxplot}
\end{layoutchg}
\end{figure}

\begin{table}[t]
\begin{layoutchg}
  \centering
  \caption{\textbf{World Model RL Large-Scale Evaluation.} We extensively evaluate WMRL across 750 episodes (seeds 1-750). WMRL is run with 15 seeds from the best BC checkpoint; \emph{best} is the best checkpoint over all seeds and \emph{avg} is the mean $\pm$ std over each seed's best checkpoint. $^\dagger$: WMRL (avg) outperforms BC under a one-sided paired $t$-test with $p<10^{-4}$.}
  \label{tab:wmrl}
    \resizebox{\linewidth}{!}{
  \begin{tabular}{lcccccc}
    \toprule
    \multirow{2}{*}{Method} & XArm & \multicolumn{2}{c}{UR10e} & \multicolumn{2}{c}{Panda} & \multirow{2}{*}{Avg} \\
    \cmidrule(lr){2-2} \cmidrule(lr){3-4} \cmidrule(lr){5-6}
     & Poke Cube & Roll Ball & PushT & Pull Cube & Pull Cube Tool & \\
    \midrule
    BC-ResNet      & 89.7 & 65.3 & 15.1 & 85.3 & 40.0 & 59.1 \\
    RLA-WMRL (avg)        & 94.2$^\dagger$\,{\scriptsize$\pm$1.1} & 65.6\,{\scriptsize$\pm$4.4} & 17.3$^\dagger$\,{\scriptsize$\pm$1.6} & 83.7\,{\scriptsize$\pm$1.7} & 37.7\,{\scriptsize$\pm$4.1} & 59.7 \\
    RLA-WMRL (best)       & \textbf{95.9} & \textbf{71.3} & \textbf{20.5} & \del{\textbf{89.7}}\add{\textbf{86.1}} & \textbf{44.7} & \del{\textbf{64.4}}\add{\textbf{63.7}} \\
    \bottomrule
  \end{tabular}}
\end{layoutchg}
\end{table}

\textbf{Evaluation and Results.}
WMRL is an offline RL method~\cite{levine2020offline} since all of the robot's actions take place inside the learned world model, without any additional interactions with the real world, which are the key cost in RL.
Therefore, independent training runs can be repeatedly performed and evaluated inside the world model for free, and the best-performing policy is retained.
This allows us to determine whether WMRL can surpass the performance of imitation learning, without using any additional interaction data from the real environment.
Specifically, we first train the BC-ResNet policy for 40 epochs, saving checkpoints at each epoch. We use a single seed (42), as BC is generally robust to initialization and batch shuffling~\cite{foster2024behavior}. Then, we apply WMRL on the BC policies. To account for variance, we run 15 trials (seeds 1-15) for 2,400 steps each, saving a checkpoint every 200 updates. 
\del{Fig.~\ref{fig:wmrl-boxplot} shows the performance distributions over 50 evaluation episodes with standard seeds (42-91). Our WMRL consistently improves the best-performing models over BC across all tasks.}
\del{Further, in addition to the standard evaluation, we conduct a large-scale evaluation across 1500 seeds (1-1500).}\add{Instead of the standard 50 evaluation episodes, we evaluate both BC and WMRL over 750 episodes (seeds 1-750).} This directly addresses a commonly known limitation in many RL works~\cite{henderson2018deep, mania2018simple}, where evaluations over a limited set of seeds can be unstable and prone to spurious results.
As shown in \add{Fig.~\ref{fig:wmrl-boxplot} and }Tab.~\ref{tab:wmrl}, \del{WMRL significantly improves policy performance on the XArm and UR10e robots.}\add{the best WMRL policy improves over BC on all five tasks (\del{+5.3\%}\addr{+4.6\%} on average), and the mean over 15 seeds is significantly higher than BC on the XArm and UR10e robots with $p<10^{-4}$ under a paired $t$-test.} Although WMRL shows lower \add{mean }performance on the Panda robot, it \del{achieves a statistically significant average gain of +1.1\% over BC across the five tasks}\add{outperforms BC when averaged over seeds and tasks (59.7\% vs. 59.1\%)}, using no additional data and only additional computation. \addr{Each WMRL run takes about 3 hours on 4 GPUs}. 
We discuss the Panda performance in Sec.~\ref{apn:limitations}. Nevertheless, one can always devise a simple meta algorithm that selectively uses WMRL depending on the task and setup, to improve BC policies without requiring any additional data, and only returns the highest-performing policies.   
We view our work as a preliminary but solid step toward rigorous standards for world model-based RL. 

\begin{wraptable}{R}{0.45\linewidth}
\begin{layoutchg}
  \centering
  \caption{\textbf{Learning from Failures.} WMRL improves PushT policy with 8{,}000 extra failed trajectories with noisy actions, which BC cannot exploit.}
  \label{tab:failure}
  \small
  \setlength{\tabcolsep}{3.5pt}
  \begin{tabular}{lcc}
    \toprule
    Method & Before & + Failure data \\
    \midrule
    BC-ResNet         & 15.1 & 13.6 \\
    RLA-WMRL (best)   & 20.5 & \textbf{22.5} \\
    RLA-WMRL (avg)    & 17.3\,{\scriptsize$\pm$1.6} & \textbf{18.5}\,{\scriptsize$\pm$2.3} \\
    \bottomrule
  \end{tabular}
\end{layoutchg}
\end{wraptable}
\leavevmode\addbegin\textbf{Learning from Failures.} Unlike BC, WMRL can learn from failed trajectories, since they still provide valid dynamics for the world model. We add 8{,}000 PushT trajectories that are executed with noisy actions and end at non-goal positions. As shown in Tab.~\ref{tab:failure}, these trajectories degrade BC\addr{, since it imitates the failed behaviors}, whereas we use them to train a better RLA-WM and further improves the success rate to 22.5\% (best) and 18.5\% (avg) through WMRL. This is a practical advantage, as failed attempts are abundant in robot learning but are typically discarded by imitation learning.\addend

\leavevmode\addbegin\textbf{Analysis.} As the policy is optimized only inside RLA-WM, it may exploit model errors. We find that RLA-WM can hallucinate successes under noisy actions, and training with noisy data reduces this gap, as shown in Fig.~\ref{fig:hallucination}, which is consistent with the gain from failure data. A moderate bound on the residual action head keeps the policy close to BC and consistently improves performance through WMRL, while a loose bound falls below BC, as in Tab.~\ref{tab:resbound}. Moreover, although VAR is time-aligned with the reference video, WMRL is not restricted to the demonstration pace. As shown in Tab.~\ref{tab:trajlen}, we find the WMRL policy solves Poke Cube and Roll Ball with significantly shorter trajectories, even with the time-synchronized reward.  Details of these studies and results are included in Appendix~\ref{apn:additional} due to space limit.\addend

\section{Limitations and Conclusion}
\label{sec:conclude}

We introduce Residual Latent Action, a compact representation of visual state dynamics, and propose RLA-WM, a state-of-the-art visual feature-based world model. Our framework enables two robot learning techniques: a minimalist world action model\add{, which also improves a VLA and a real-robot policy,} and a visual RL framework in RLA-WM.  We believe our work is a strong step forward for visual feature-based world models. 
However, some limitations are still worth discussing: (1) Task-irrelevant background motion 
can cause visual changes between $s_t$ and $s_{t+h}$, which are common for humanoids or eye-in-hand cameras. Forcing RLA to encode such information could waste representation capacity and degrade the quality of RLA. \add{Although RLA remains effective with a moving wrist camera (Sec.~\ref{sec:learn-from-actionless}), large camera motion is still challenging.} A solution is to learn view-independent RLA in the 3D space; (2) The visual changes between $s_{t+h}$ and $s_t$ can result from occlusion or partial observation, such as when the viewpoint changes, or an object disappears and later reappears. Understanding these motions requires reasoning over historical observations, which is difficult to capture from only a single frame pair $(s_t, s_{t+h})$. A promising solution is to extend RLA to multiple frames; (3) Our evaluation focuses on small-scale datasets and simulation\add{, with one real-robot task}. In doing so, we ensure reproducibility and isolate our method-driven gains from mere data scaling. While this demonstrates the efficacy of our method, adapting our framework to internet-scale data in an open-world setting remains an open question and promising direction.
\add{(4) WMRL relies on reference videos to compute VAR, and RLA-WM may still hallucinate successes for out-of-distribution actions. Learning rewards without aligned demonstrations and building conservative world models that are aware of their own uncertainty are promising future directions.}

\appendix
\clearpage
\newpage
\section{Appendix}
\label{sec:apn}
\renewcommand{\thefigure}{A\arabic{figure}}
\setcounter{figure}{0}
\renewcommand{\thetable}{A\arabic{table}}
\setcounter{table}{0}


\subsection{Code}

The source code of our work is included in the supplementary folder \texttt{Code}. 
Please see \texttt{Code/README.md} for instructions on installation, dataset setup, downloading pre-trained models, running the demo Jupyter notebook, and training.

\subsection{Implementation Details}
\label{apn:details}

Here we summarize the details and hyperparameters for each section of our experiments. Note that we use DINOv3-Large with channel size 1024, and AdamW optimizer~\cite{loshchilovdecoupled} for all experiments. 

Our RLA autoencoder training code is built on TRELLIS~\cite{xiang2025structured}, which is released under the MIT License. Our imitation learning code is built on Diffusion Policy~\cite{chi2025diffusion}, also released under the MIT License. We use pre-trained DINOv3 models~\cite{simeoni2025dinov3}, which are released under the DINOv3 License. We use the ManiSkill rigid-body simulation suite~\cite{mu2maniskill} (Apache 2.0 License; assets under CC BY-NC 4.0) and the IWS dataset~\cite{wang2026interactive} (MIT License).

\textbf{RLA Autoencoder.} We use 12 self-attention layers for both $f_\text{enc}$ and $f_\text{dec}$, with 16 heads and a channel size of 1024. Input images are $512\times512$. During training, we randomly sample frame pairs within a horizon of 200 for the IWS dataset and 100 for ManiSkill. We use a batch size of 128, a learning rate of $10^{-4}$, and train for 100k steps. Reconstruction of $s_{t+h}$ uses both L1 and MSE loss, each with weight 1.0. On ManiSkill, we bias frame sampling toward object movement: with probability 0.9 we sample a frame pair containing object movement (object movements are pre-recorded in the simulator), otherwise we sample a random frame. On IWS (a real-world dataset), we sample frames uniformly at random. Unless specified otherwise, all experiments use an RLA size $|z| = 2048$. Specifically, the encoder output uses 32 query tokens, each projected to dimension 64. To render DINO tokens back to images, we train a separate decoder-only UNet with 4 upsampling deconvolution blocks; each block doubles the spatial resolution and halves the channel dimension. 

\textbf{RLA World Model.} The condition network uses 8 self-attention layers with 16 heads, channel size 1024, and 32 query tokens. The flow matching network also uses 8 self-attention layers and channel size 1024, but operates on a smaller token count: only 64 tokens total (32 from the condition network and 32 for the noisy RLA $z_\tau$), and its output dimension is 64, consistent with our RLA representation where $|z| = 2048 = 32 \times 64$. We use a maximum action horizon of 15. The RLA-WM is trained for 100k steps with learning rate $10^{-4}$ and batch size 64. During inference, we use 30 Euler ODE steps for flow matching. Different robots have different action sizes due to their kinematics: for ManiSkill, Panda (8), UR10 (5), XArm-Robotiq (12); for the IWS ALOHA robot, we use the dataset's provided actions --- the rope task has action size 14 (7 joints for each ALOHA arm), the box task has action size 8 (3D Cartesian coordinates of each gripper, plus an additional dimension to control gripper openness), the Push-T task has has action size 4 (2D table coordinates for each of two arms). Before feeding actions into the condition network, we embed them via a robot-specific MLP: actions are first padded to the maximum horizon (e.g., for Panda with horizon 15, we pad to shape $8 \times 15$), then passed through the MLP. RLA-WM and its autoencoder each take 3 days on 4$\times$ A6000 GPUs (48GB) with 256GB RAM. The dynamics component of RLA-WM is trained per robot on ManiSkill, per scene on IWS, while the RLA autoencoder is trained per dataset.

\textbf{Learning from Actionless Videos with RLA.} Input images are $128 \times 128$, and we use an action chunk size of 12 during training and pick the first 8 to execute each step during inference. The visual token self-attention uses 6 layers with channel size 256 and 8 heads. Training runs for 40 epochs with batch size 64, learning rate $3 \times 10^{-4}$, weight decay $10^{-4}$, and a cosine learning rate scheduler. For the BC-ResNet baseline, which uses only 5\% of videos (those with actions), the epoch size is much smaller; we therefore scale up the number of epochs to match the total iterations of latent action-based training while keeping the same evaluation and checkpoint frequency (40 times). We save a checkpoint and evaluate after each epoch. Evaluation runs 50 episodes (random seeds 42-91), each with a maximum of 100 steps. Each policy is trained per task. A single training trial takes one day on an A4500 Ada GPU (24GB).

\begin{figure}[tp]
    \centering
    \includegraphics[width=\linewidth]{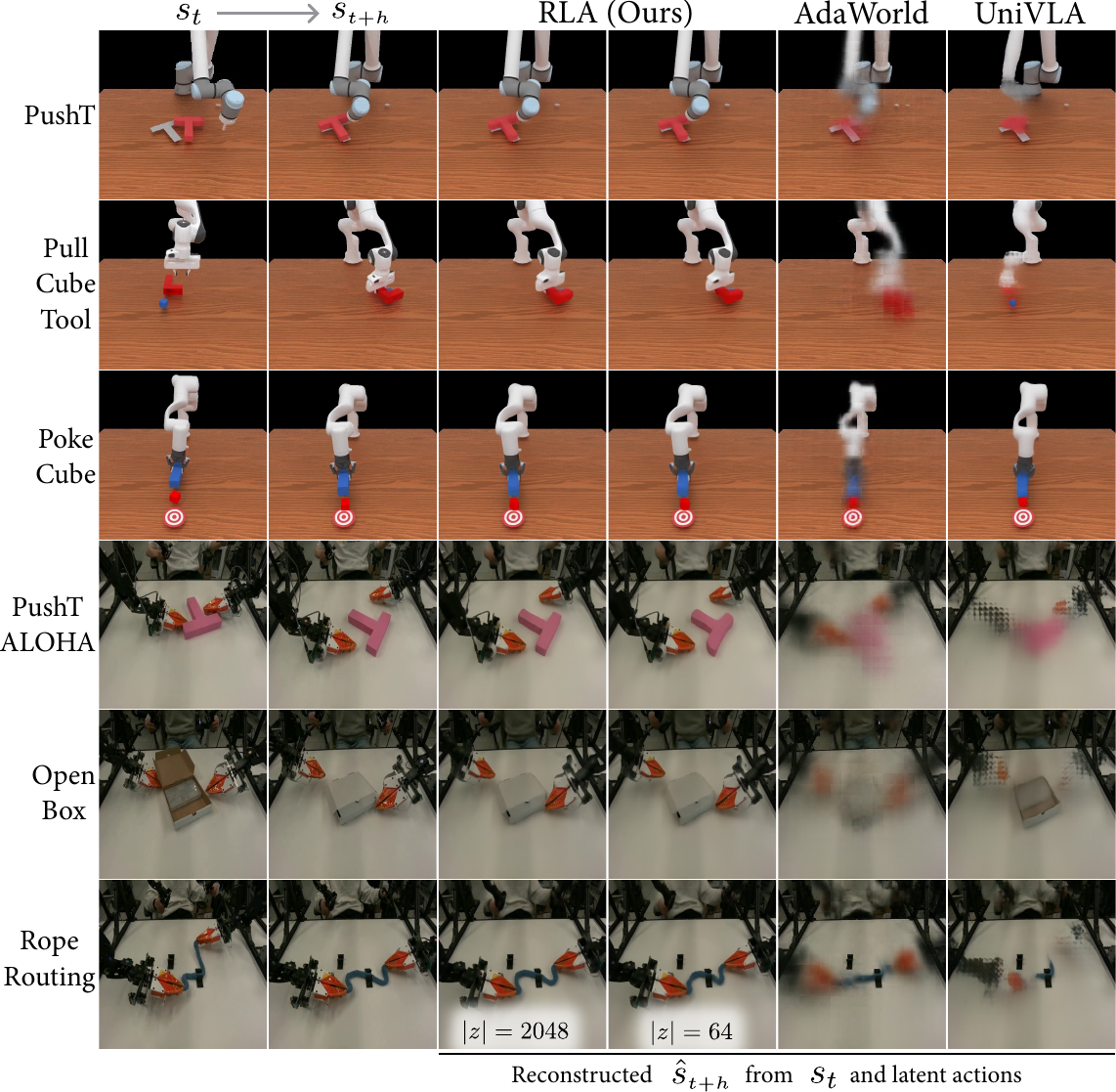}
    \caption{\textbf{Reconstruction Quality Comparison.} 
    We compare the reconstruction $\hat{s}_{t+h}$ decoded by $f_{\text{dec}}$ from the current frame $s_t$ and a latent action $z$, where $z$ is encoded by $f_{\text{enc}}$ from the pair $(s_t, s_{t+h})$. RLA produces accurate reconstructions even with a compact latent dimension $|z| = 64$. This is particularly impressive  given the $1024 \times 1024$ dimension of DINO token $s_{t+h}$. In contrast, AdaWorld~\cite{gao2025adaworld} and UniVLA~\cite{bu2025univla} yield severely blurred and inaccurate predictions with latent dimensions 2048 and 256, respectively. This demonstrates that RLA captures sufficient predictive information for dynamics and enables accurate future token decoding in a single feedforward pass.}
    \label{fig:rla-predictive}
\end{figure}

\textbf{Visual RL within RLA-WM.} We inject LoRA adapters into all linear and convolutional layers of the pre-trained BC ResNet policy. Unlike the architecture used in our  learning from actionless video experiments, this policy does not use self-attention over spatial tokens; instead, we apply global average pooling followed directly by an MLP for action prediction. The residual action head is a 3-layer MLP with hidden dimensions of 512 and 256. To stabilize exploration, the residual mean and standard deviation are bounded to 0.1 radians (joint angle) using $\tanh$ and softplus activations, respectively\add{, which we ablate in Sec.~\ref{apn:additional}}. For the Video Aligned Reward, the \texttt{Poke Cube} task uses the final goal frame $s_T$, while all other tasks compute the reward against the time-synchronized frame $s_t$. To augment initial state diversity, we initialize episodes from a random intermediate video frame with a $0.5$ probability, controlled deterministically via the random seed. We apply a reward scale of 5 and rely exclusively on PPO gradients, without any terminal rewards and auxiliary behavior cloning losses.

RL optimization is conducted over 15 random seeds using a discount factor $\gamma = 0.9$, a GAE~\cite{schulman2015high} parameter $\lambda = 0.95$, and a learning rate of $1 \times 10^{-4}$. We vector-parallelize 112 RLA-WM environments across multi-GPU setups. To ensure reproducibility, random seeds are uniquely mapped to the environment index rather than the GPU worker, so the same seed yields similar training results regardless of the number of GPUs. We rollout 300 steps inside the world model, each with 4 action-chunking steps of chunk size 8, and update the policy 8 times with a PPO batch size of 224. Every 200 updates we evaluate and save a checkpoint. 
Each training trial takes 3 hours on a 4-GPU A6000 machine or a 7-GPU A4500 Ada machine.
We do not seed the Euler ODE steps for flow matching; although theoretically stochastic, we observe negligible effect on results (unlike environment resets or initial frame sampling). For final evaluation to select the best model for BC or RL, we run \del{1500}\add{750} evaluation episodes with seeds \del{1-1500}\add{1-750}.

\begin{figure}[t]
    \centering
    \includegraphics[width=\linewidth]{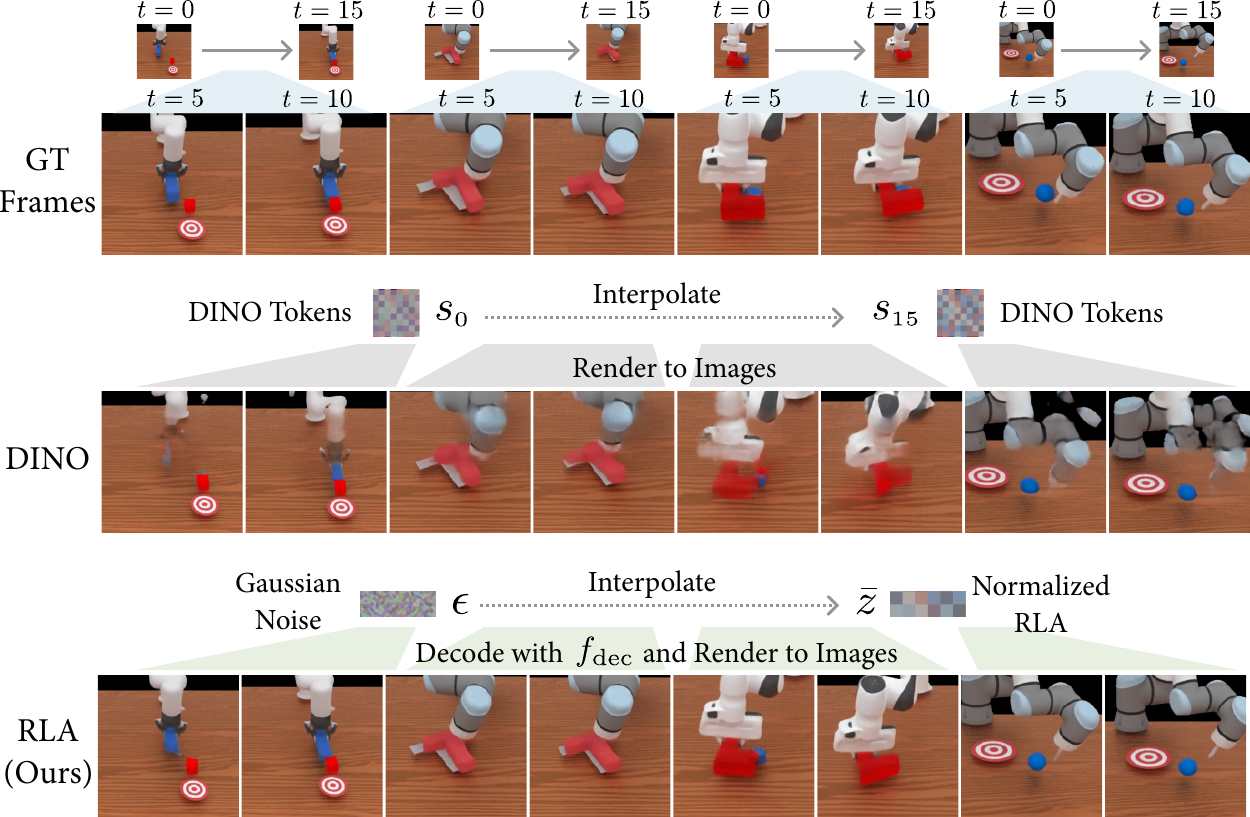}
    \caption{\textbf{RLA Temporal Topology.} We encode RLA $z$ from $(s_t, s_{t+h})$ and normalize it to $\bar{z} = (z - \mu)/\sigma$, where $\mu$ and $\sigma$ are estimated from data. We then interpolate between Gaussian noise $\epsilon$ and $\bar{z}$, then denormalize and decode the result. For example, $(\epsilon + \bar{z})/2$ yields a reconstruction that approximates $s_{t+h/2}$. This indicates that the RLA latent space inherently captures temporal progression. In contrast, interpolating DINO tokens between $s_t$ and $s_{t+h}$ produces inferior results. }
    \label{fig:rla-topology}
\end{figure}

To bridge the \textit{neural-to-sim} gap --- the world model renders images from DINO tokens that differ slightly from simulator ray-traced images --- we extract DINO tokens from each observed image, decode RGB images using our pre-trained UNet, and use the decoded image as observation. This largely removes the gap and is applied during both BC pre-training and evaluation. We acknowledge that this increases computational cost and may lower the performance ceiling, but we believe scaling the world model and policy will improve robustness to image styles in the future.

\subsection{\texorpdfstring{\add{Additional Experiments and Analyses}}{Additional Experiments and Analyses}}
\label{apn:additional}

\add{We provide additional experiments on the temporal topology of RLA, RLA for VLAs and real robots, and analyses of WMRL on model exploitation and trajectory length.}

\add{\textbf{Temporal Topology of RLA.} We quantify the temporal topology of RLA (Fig.~\ref{fig:rla-topology}) on 2,000 trajectory segments with horizon $h=15$. For each segment, we encode RLA $z$ from $(s_t, s_{t+h})$, interpolate between the normalized $\bar{z}$ and Gaussian noise $\epsilon$ with 11 evenly spaced coefficients over 5 noise draws, and decode each interpolant. We then match each decoded frame to the closest real frame within $s_{t:t+h}$, and report the Spearman correlation between the matched and expected frame indices, as well as the mean frame index error. As controls, we linearly blend $s_t$ and $s_{t+h}$ in DINO token space and in pixel space. As shown in Tab.~\ref{tab:interp}, RLA interpolation yields the strongest temporal correlation (0.97) and the lowest frame index error (1.88 out of 15 frames). The same trend holds for $h=30$, where RLA achieves a Spearman correlation of 0.97 (vs.\ 0.92 for DINO and 0.87 for pixel blending) and a frame index error of 4.6 (vs.\ 5.9 and 6.8). This temporal structure emerges from RLA despite that RLA autoencoder is only trained on frame pairs.}

\begin{table}[t]
\begin{layoutchg}
  \centering
  \caption{\textbf{Temporal Correlation of Latent Interpolation.} We quantify the temporal topology of RLA (Fig.~\ref{fig:rla-topology}) on 2,000 trajectory segments whose start and end frames are 15 steps apart. We interpolate between the normalized RLA and Gaussian noise, decode each interpolant, and match it to the closest real intermediate frame. As baselines, we linearly blend the start and end frames in pixel space and in DINO token space. We report the Spearman correlation~\cite{spearman1961proof} between the matched and expected frame indices, and the mean frame index error. RLA interpolation follows the temporal order most closely.}
  \label{tab:interp}
  \begin{tabular}{lccc}
    \toprule
     & Pixel Blend & DINO Blend & RLA Interp.\ (Ours) \\
    \midrule
    Spearman $\uparrow$            & 0.87 & 0.93 & \textbf{0.97} \\
    Frame Index Error $\downarrow$ & 3.13 & 2.51 & \textbf{1.88} \\
    \bottomrule
  \end{tabular}
\end{layoutchg}
\end{table}

\begin{figure}[t]
\begin{layoutchg}
  \centering
  \begin{subfigure}[b]{0.345\linewidth}
    \centering
    \includegraphics[width=\linewidth]{figures/rebuttal/preview-task.pdf}
    \caption{Task Setups}
    \label{fig:preview-task}
  \end{subfigure}\hfill
  \begin{subfigure}[b]{0.625\linewidth}
    \centering
    \includegraphics[width=\linewidth]{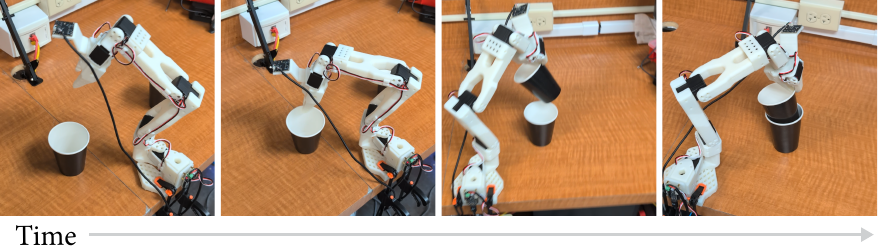}
    \caption{Real-Robot Rollout}
    \label{fig:realrobot}
  \end{subfigure}
  \\[0.8em]
  \begin{subfigure}[b]{0.44\linewidth}
    \centering
    \includegraphics[width=\linewidth]{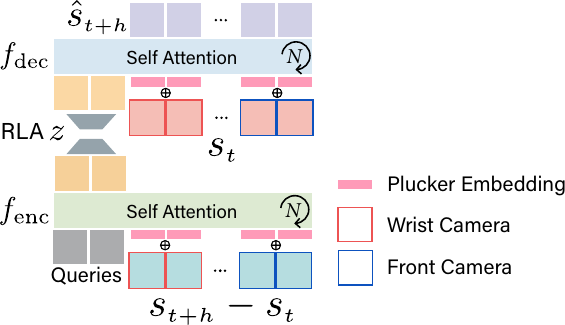}
    \caption{Multi-View RLA Autoencoder}
    \label{fig:rla-multiview}
  \end{subfigure}\hfill
  \begin{subfigure}[b]{0.52\linewidth}
    \centering
    \includegraphics[width=\linewidth]{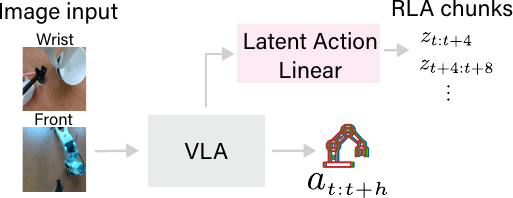}
    \caption{Policy with RLA Chunk Prediction}
    \label{fig:real-robot-arch}
  \end{subfigure}
  \caption{\textbf{RLA for VLAs and Real Robots.} (a) We evaluate RLA on the LIBERO benchmark with VLA-Adapter (Tab.~\ref{tab:libero}) and on real-world cup stacking with an SO-101 robot (Tab.~\ref{tab:real}). (b) We show a real-robot rollout, where the policy picks up a cup and stacks it onto the other cup. (c) Since both tasks use a front camera and a moving wrist camera, we extend the RLA autoencoder to multiple views. We concatenate the DINO tokens of both cameras, add a Pl\"ucker embedding~\cite{zhang2024cameras} to the tokens of each camera, and encode their residuals into a single RLA $z$. (d) The policy takes the wrist and front images as input, and predicts the robot actions $a_{t:t+h}$ and, through an additional linear head, the RLA chunks $z_{t:t+4}, z_{t+4:t+8}, \dots, z_{t+h-4:t+h}$, where $z_{t:t+4}$ is the RLA encoded from $(s_t, s_{t+4})$. The RLA head serves as an auxiliary training target and adds no inference cost.}
  \label{fig:rla-vla-real}
\end{layoutchg}
\end{figure}

\begin{figure}[t]
\begin{layoutchg}
  \begin{minipage}[b]{0.6\linewidth}
    \centering
    \definecolor{hgt}{RGB}{51,51,51}%
    \definecolor{hwm}{RGB}{214,96,88}%
    \definecolor{hnoisy}{RGB}{58,124,192}%
    \newcommand{\hpt}[4][]{\node[font=\scriptsize\bfseries, text=#2, inner sep=0.5pt, #1]
      at ({(47.9+27.84*#3)/341}, {1-(159.2-1.245*#4)/188}) {#4};}%
    \begin{tikzpicture}
      \node[anchor=south west, inner sep=0] (img) at (0,0)
        {\includegraphics[width=\linewidth]{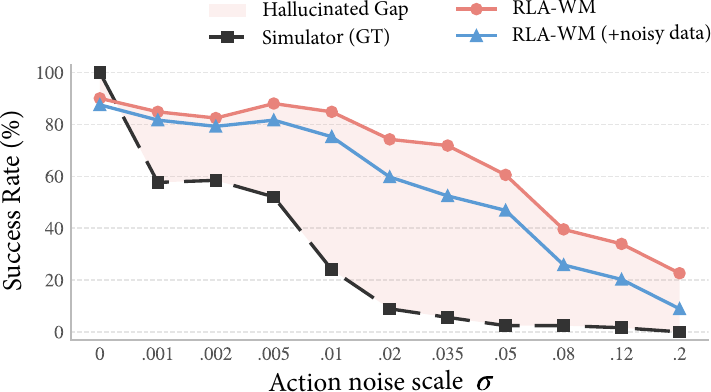}};
      \begin{scope}[x={(img.south east)}, y={(img.north west)}]
        \hpt[above=2.5pt]{hgt}{0}{100}
        \hpt[above right=1pt, xshift=4pt]{hwm}{0}{90.0}
        \hpt[below=3pt]{hnoisy}{0}{87.5}
        \hpt[above right=1.5pt]{hwm}{4}{84.8}
        \hpt[below left=1.5pt]{hnoisy}{4}{75.2}
        \hpt[above right=1.5pt]{hgt}{4}{24.0}
        \hpt[right=3pt, overlay]{hwm}{10}{22.6}
        \hpt[right=3pt, overlay]{hnoisy}{10}{8.9}
        \hpt[right=3pt, overlay]{hgt}{10}{0.0}
      \end{scope}
    \end{tikzpicture}
    \caption{\textbf{Hallucinated Successes under Noisy Actions.} We replay successful PushT trajectories with drifting action noise~\cite{lillicrap2015continuous} of scale $\sigma$, both in the simulator and in RLA-WM, and report the success rates (\%). The shaded area is the hallucinated gap, i.e., successes predicted by RLA-WM that do not occur in the simulator. RLA-WM predicts many more successes than the simulator under moderate noise, and training it with additional noisy data narrows this gap.}
    \label{fig:hallucination}
  \end{minipage}\hfill
  \begin{minipage}[b]{0.37\linewidth}
    \centering
    \captionof{table}{\textbf{Residual Bound and WMRL on PushT.} The residual action head outputs a mean offset $\delta_\mu=b\tanh(\cdot)$ on top of the BC policy, where $b=0$ recovers BC and ``Unbounded'' removes both $\tanh$ and the bound. We report success rates (\%) over 50 episodes as the mean $\pm$ std of each seed's best checkpoint over 5 seeds. A moderate bound consistently improves over BC.}
    \label{tab:resbound}
    \small
    \begin{tabular}[b]{lc}
      \toprule
      Residual bound $b$ & Success Rate \\
      \midrule
      0 (BC)    & 18.0 \\
      0.05      & \textbf{22.8}\,{\scriptsize$\pm$1.8} \\
      0.1       & 22.0\,{\scriptsize$\pm$1.4} \\
      0.25      & 14.0\,{\scriptsize$\pm$3.7} \\
      Unbounded & 22.7\,{\scriptsize$\pm$2.3} \\
      \bottomrule
    \end{tabular}
  \end{minipage}
  \par\vspace{1.2em}
  \centering
  \captionof{table}{\textbf{Trajectory Length of Successful Rollouts.} We report the mean number of control steps on the evaluation episodes that \emph{both} BC and WMRL solve. Numbers in parentheses are the differences from BC, and $^\dagger$ indicates that WMRL is shorter in at least 14 of the 15 seeds under Wilcoxon test of $p<10^{-4}$. WMRL solves Poke Cube and Roll Ball significantly faster than BC.}
  \label{tab:trajlen}
  \small
  \begin{tabular}{lccccc}
    \toprule
    \multirow{2}{*}{Method} & XArm & \multicolumn{2}{c}{UR10e} & \multicolumn{2}{c}{Panda} \\
    \cmidrule(lr){2-2} \cmidrule(lr){3-4} \cmidrule(lr){5-6}
     & Poke Cube & Roll Ball & PushT & Pull Cube & Pull Cube Tool \\
    \midrule
    BC-ResNet & 42.6 & 30.1 & 35.9 & 24.6 & 53.2 \\
    RLA-WMRL  & \textbf{34.7}$^\dagger$ {\scriptsize($-$7.9)} & \textbf{23.8}$^\dagger$ {\scriptsize($-$6.3)} & 36.8 & 25.9 & \textbf{51.8} {\scriptsize($-$1.4)} \\
    \bottomrule
  \end{tabular}
\end{layoutchg}
\end{figure}

\add{\textbf{Hallucinated Successes.} Since WMRL optimizes the policy only inside RLA-WM, the policy may exploit model errors. To study this, we replay successful PushT trajectories with drifting Ornstein--Uhlenbeck action noise of scale $\sigma$, both in the simulator and in RLA-WM, and compare their success rates (Fig.~\ref{fig:hallucination}). Without noise, RLA-WM reproduces most successes (90.0\%). As $\sigma$ increases, the success rate in the simulator quickly drops (24.0\% at $\sigma=0.01$), while RLA-WM still predicts 84.8\%. That is, RLA-WM hallucinates successes for out-of-distribution actions, which the policy could exploit during RL. Training RLA-WM with additional noisy trajectories narrows this gap for all $\sigma>0$ (e.g., from 22.6\% to 8.9\% at $\sigma=0.2$), which is consistent with the improvement of WMRL from failure data (Tab.~\ref{tab:failure}). Conservatism in world models remains an open problem, and we view this analysis as a first step.}

\add{\textbf{Residual Bound.} The residual action head outputs the mean offset $\delta_\mu = b\tanh(\cdot)$ on top of the BC policy, which bounds how far the policy can move away from BC. We vary $b$ on PushT and run WMRL with 5 seeds (3 for unbounded). As shown in Tab.~\ref{tab:resbound}, a moderate bound ($b\in\{0.05, 0.1\}$) consistently improves over BC with low variance, while a loose bound ($b=0.25$) falls below BC, likely because the policy drifts away from the BC prior and exploits model errors. Removing the bound and $\tanh$ altogether also performs well on PushT, but with a larger variance. We use $b=0.1$ in all experiments.}

\add{\textbf{Trajectory Length.} VAR compares the predicted tokens with the time-synchronized reference $s_t$ (or the goal $s_T$ for Poke Cube), which may seem to restrict the policy to the pace of the demonstrations. To verify this, we compare the number of control steps of BC and WMRL on the evaluation episodes that both solve, paired by evaluation seed (Tab.~\ref{tab:trajlen}). WMRL solves Poke Cube and Roll Ball with significantly shorter trajectories ($-7.9$ and $-6.3$ steps; shorter in at least 14 of 15 seeds, Wilcoxon $p<10^{-4}$), and is comparable to BC on the other tasks. Notably, Roll Ball uses the time-synchronized reward, showing that VAR does not force the policy to track the demonstration speed. }


\subsection{Limitations and Future Directions}
\label{apn:limitations}

\begin{figure}[t]
    \centering
    \includegraphics[width=\linewidth]{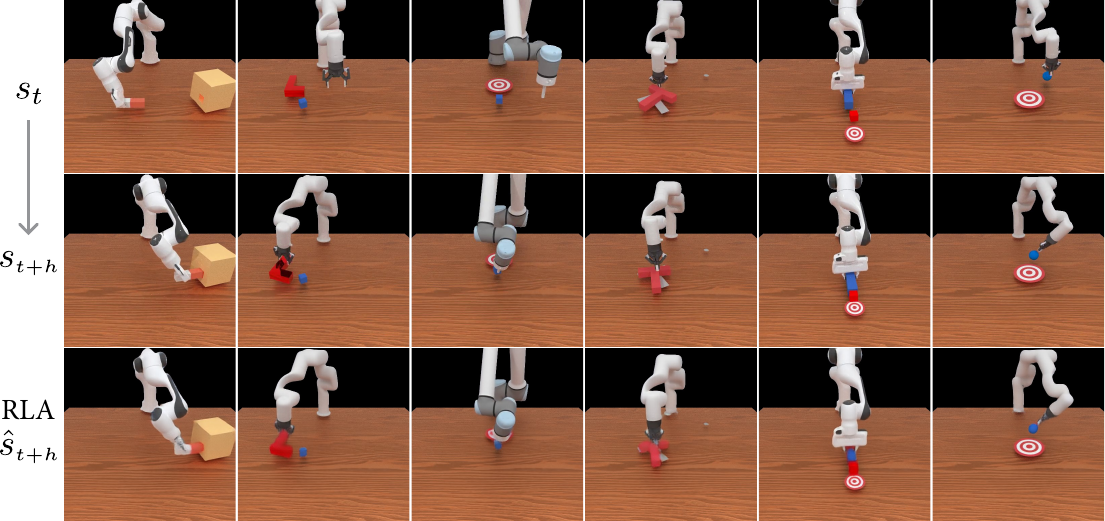}
    \caption{\textbf{RLA Generalization on Novel Tasks.}  We apply the pre-trained RLA autoencoder to a previously unseen setup. For example, the original Panda robot is replaced with an XArm robot with a Robotiq gripper for the \texttt{Pull Cube with Tool} task. This interaction type was never observed during RLA autoencoder training. Yet, RLA maintains high reconstruction fidelity. 
    Notably, this cross-embodiment generalization is achieved by training solely on the limited ManiSkill dataset, without relying on large-scale video pre-training of the RLA autoencoder. }
    \label{fig:rla-generalizable}
\end{figure}

We summarize four key limitations and corresponding future directions. 
\begin{enumerate}
    \item \textbf{Background and random motion.} Our RLA is learned from residuals between pairs of DINO tokens $(s_t, s_{t+h})$. However, task-irrelevant background motion or workspace randomness (e.g., in humanoid robot or eye-in-hand camera) can also cause visual changes between $s_t$ and $s_{t+h}$. Learning to encode those randomness-driven motions could waste representation capacity and degrade the RLA latent space. A promising fix is to move from 2D image learning to 3D, projecting DINO tokens into 3D~\cite{xiang2025structured} and learning view-independent 3D RLA. 

    \item \textbf{Memory and partial observability.} Our RLA-WM predicts $s_{t+h}$ from $s_t$ and $a_{t:t+h}$, yet changes may depend on $s_{<t}$ due to occlusion (e.g., an object disappears and reappears). Because RLA $z$ is learned from a single frame pair, it must memorize the object in the latent space rather than encoding true movement and occlusion events. Extending RLA to condition on multiple frames is a natural solution.

    \item \textbf{Proprioceptive world model.} Our RLA-WM predicts only visual state evolution via RLA, but not future proprioceptive states.  Proprioception input has been shown to be useful for policy learning. Extending the world model to predict both would broaden applicability. 

    \item \textbf{Scaling to larger datasets.}  We deliberately evaluated on small-scale ManiSkill and IWS datasets to isolate method-driven gains from mere data scaling --- many prior works scale first and leave it unclear whether improvements come from data volume or the method itself. Our clear, reproducible results on small data demonstrate the core properties of RLA and RLA-WM. Therefore, scaling to massive real-world datasets is a promising future step.
\end{enumerate}

\textbf{Panda WMRL Results.} WMRL underperforms BC on the Panda robot for the \texttt{Pull Cube} and \texttt{Pull Cube with Tool} tasks. We train the dynamics component of  RLA-WM per robot, yet only the Panda results show a consistent drop. We attribute this to a combination of factors: (1) the Panda's kinematic structure, (2) the camera viewpoint, and (3) insufficient action diversity in the demonstration data.

The Panda arm has 8 degrees of freedom (DoF), whereas the XArm has 7 (the Robotiq gripper consists of 6 correlated joints that provide 1 effective DoF), and the UR10e has 5 (the sixth joint is ineffective due to a cylindrical end-effector). Since our policy predicts future joint angles, higher-dimensional action spaces naturally require more data. However, all robots receive the same number of demonstrations. Moreover, the two tasks of Panda both involve pulling a cube toward the robot's side of the table, producing action patterns with limited diversity. This pulling motion frequently causes occlusion from our front top-view camera, hindering the world model's ability to capture the mapping from joint angles to visual changes accurately. Additionally, the Panda's gripper fingers are small, making it difficult for the world model to capture fine visual details. We believe that adding multiple camera views, increasing task variety (and thus data diversity), and collecting more demonstrations overall would resolve these issues.

\begin{figure}[h]
    \centering
    \begin{subfigure}[b]{\textwidth}
        \centering
        \includegraphics[width=\textwidth]{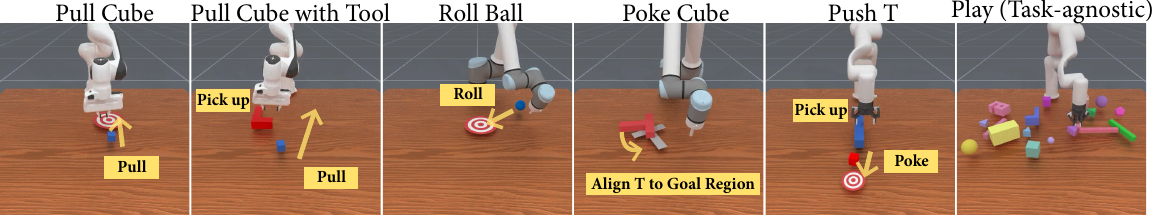} 
        \caption{ManiSkill Environment~\cite{mu2maniskill}}
        \label{fig:sub1}
    \end{subfigure} \\[1em]
    \begin{subfigure}[b]{0.6\textwidth}
        \centering
        \includegraphics[width=\textwidth]{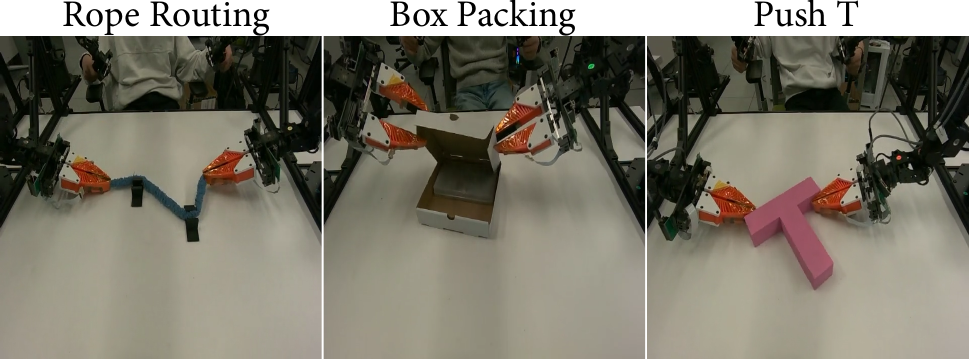} %
        \caption{IWS Dataset~\cite{wang2026interactive}}
        \label{fig:sub2}
    \end{subfigure}
    \caption{\textbf{Overview of Tasks and Datasets.} ManiSkill includes five tasks across three robots: Panda --- \texttt{Pull Cube} (pull cube to target area) and \texttt{Pull Cube with Tool} (grasp L-shaped tool to hook cube within a distance to the robot); UR10 with cylinder end-effector — \texttt{Roll Ball} (move ball to target region) and \texttt{Push T} (align T-shaped object with T-shaped goal area); XArm with Robotiq gripper — \texttt{Poke Cube} (pick blue stick and poke cube to target). IWS (ALOHA robot) includes \texttt{Rope Routing} (route rope around marked anchors on table), \texttt{Box Packing} (open, close, or move the box), and \texttt{Push T} (continuous two-arm interaction with the T-object to create diverse movements). }
    \label{fig:env}
\end{figure}


\begin{table}
    \centering
    \caption{Detailed Per-Robot Evaluation of Future Frame Prediction on ManiSkill. Our RLA-WM achieves the best results across all robots and metrics. }
    \label{tab:maniskill-detailed}
    \resizebox{\textwidth}{!}{
    \begin{tabular}{lccccccccc}
        \toprule
        \multirow{2}{*}{Model} & \multicolumn{3}{c}{LPIPS $\downarrow$} & \multicolumn{3}{c}{SSIM $\uparrow$} & \multicolumn{3}{c}{DINO L1 $\downarrow$} \\
        \cmidrule(lr){2-4} \cmidrule(lr){5-7} \cmidrule(lr){8-10}
        & Panda & XArm & UR10 & Panda & XArm & UR10 & Panda & XArm & UR10 \\
        \midrule
        DINO-WM & 0.187 & 0.120 & 0.163 & 0.841 & 0.896 & 0.857 & 0.098 & 0.056 & 0.079 \\
        RAE & 0.310 & 0.313 & 0.349 & 0.719 & 0.730 & 0.701 & 0.139 & 0.166 & 0.124 \\
        FM-WM & 0.147 & 0.119 & 0.114 & 0.872 & 0.900 & 0.898 & 0.075 & 0.059 & 0.054 \\
        Vid2World & 0.219 & 0.186 & 0.192 & 0.685 & 0.720 & 0.710 & 0.094 & 0.081 & 0.077 \\
        \midrule
        RLA-WM (Ours) & \textbf{0.091} & \textbf{0.061} & \textbf{0.061} & \textbf{0.912} & \textbf{0.941} & \textbf{0.941} & \textbf{0.035} & \textbf{0.028} & \textbf{0.027} \\
        \bottomrule
    \end{tabular}
    }
\end{table}

\begin{table}
    \centering
    \caption{Detailed Per-Task Evaluation of Future Frame Prediction on IWS. Our RLA-WM achieves the best results on all tasks except rope routing, where it performs within a small margin of the best.}
    \label{tab:iws-detailed}
    \resizebox{\textwidth}{!}{
    \begin{tabular}{lccccccccc}
        \toprule
        \multirow{2}{*}{Model} & \multicolumn{3}{c}{LPIPS $\downarrow$} & \multicolumn{3}{c}{SSIM $\uparrow$} & \multicolumn{3}{c}{DINO L1 $\downarrow$} \\
        \cmidrule(lr){2-4} \cmidrule(lr){5-7} \cmidrule(lr){8-10}
        & Box & PushT & Rope & Box & PushT & Rope & Box & PushT & Rope \\
        \midrule
        DINO-WM & 0.223 & 0.279 & \textbf{0.167} & 0.830 & 0.774 & \textbf{0.870} & 0.055 & 0.077 & \textbf{0.043} \\
        RAE & 0.596 & 0.523 & 0.530 & 0.549 & 0.669 & 0.656 & 0.167 & 0.153 & 0.157 \\
        FM-WM & 0.347 & 0.373 & 0.359 & 0.750 & 0.735 & 0.738 & 0.115 & 0.121 & 0.119 \\
        Vid2World & 0.384 & 0.398 & 0.382 & 0.716 & 0.702 & 0.714 & 0.134 & 0.142 & 0.142 \\
        \midrule
        RLA-WM (Ours) & \textbf{0.206} & \textbf{0.204} & 0.177 & \textbf{0.839} & \textbf{0.845} & 0.857 & \textbf{0.055} & \textbf{0.055} & 0.048 \\
        \bottomrule
    \end{tabular}
    }
\end{table}

\begin{figure}[h]
    \centering
     \makebox[\linewidth][l]{%
        \hspace*{-1.5em}
    \includegraphics[width=\linewidth]{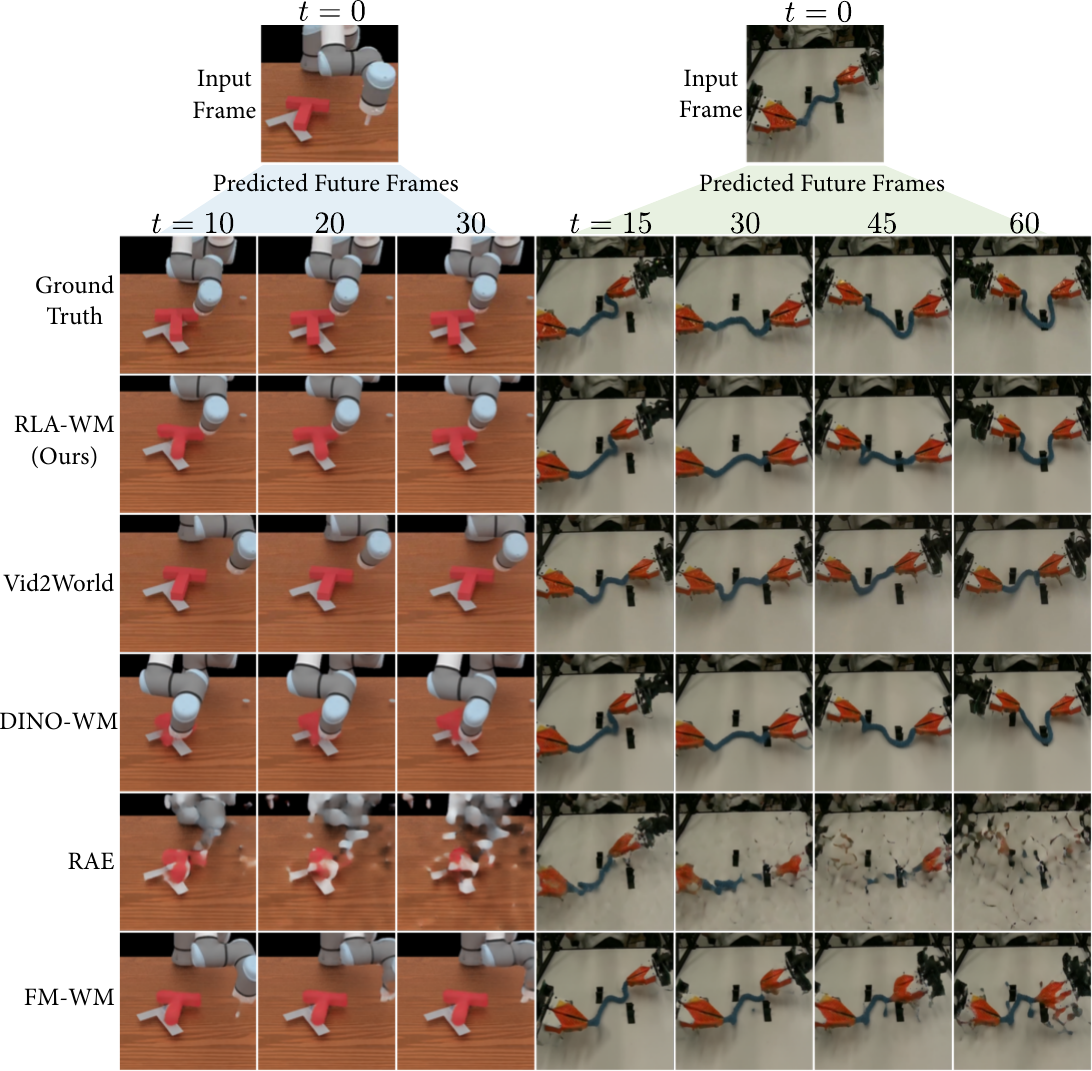}}
    \caption{\textbf{Additional Qualitative Comparison for RLA-WM.}}
    \label{fig:wm-result-1}
\end{figure}

\begin{figure}[h]
    \centering
     \makebox[\linewidth][l]{%
        \hspace*{-1.5em}
    \includegraphics[width=\linewidth]{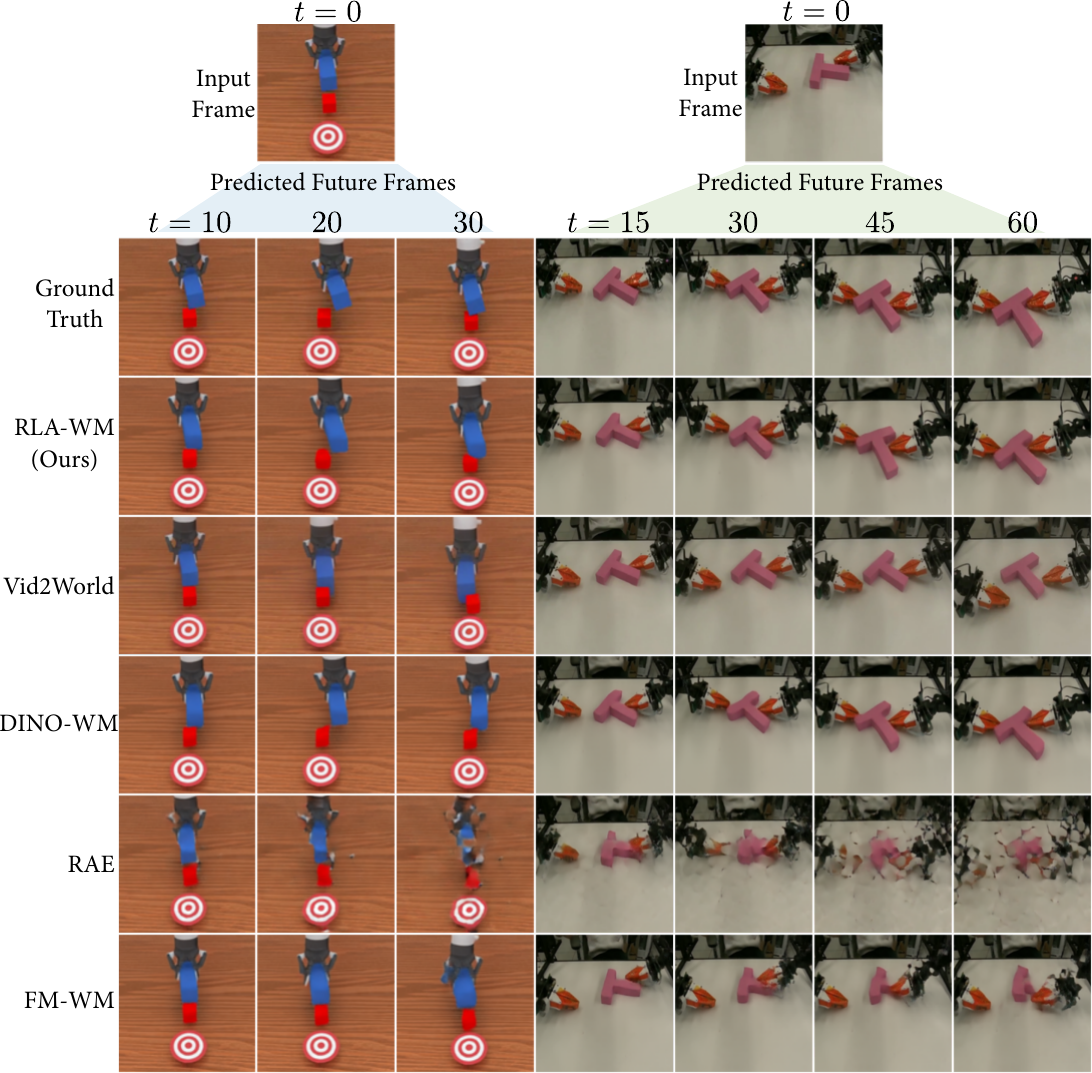}}
    \caption{\textbf{Additional Qualitative Comparison for RLA-WM.}}
    \label{fig:wm-result-2}
\end{figure}

\begin{figure}[h]
    \centering
     \makebox[\linewidth][l]{%
        \hspace*{-1.5em}
    \includegraphics[width=\linewidth]{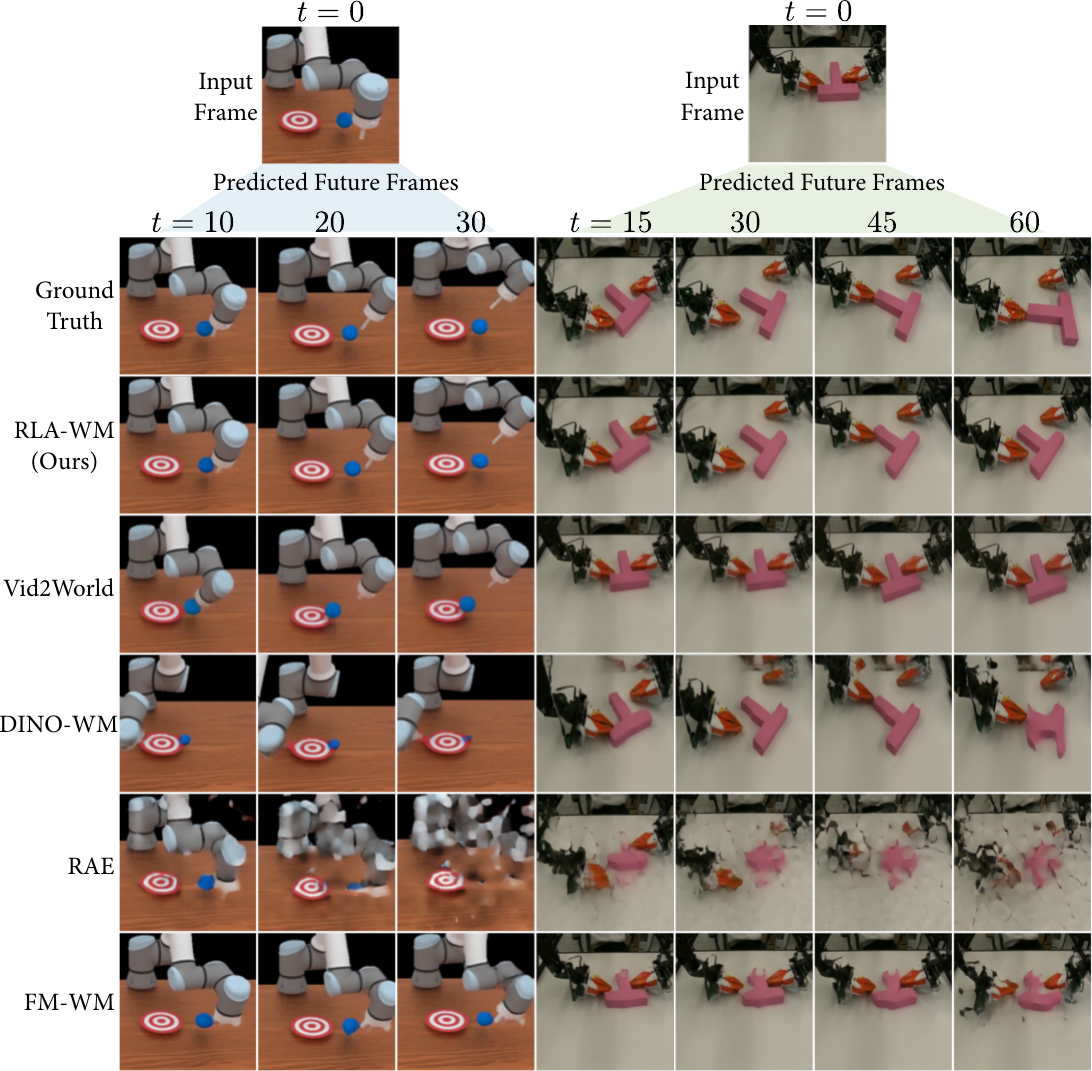}}
    \caption{\textbf{Additional Qualitative Comparison for RLA-WM.}}
    \label{fig:wm-result-3}
\end{figure}

\begin{figure}[h]
    \centering
     \makebox[\linewidth][l]{%
        \hspace*{-1.5em}
    \includegraphics[width=\linewidth]{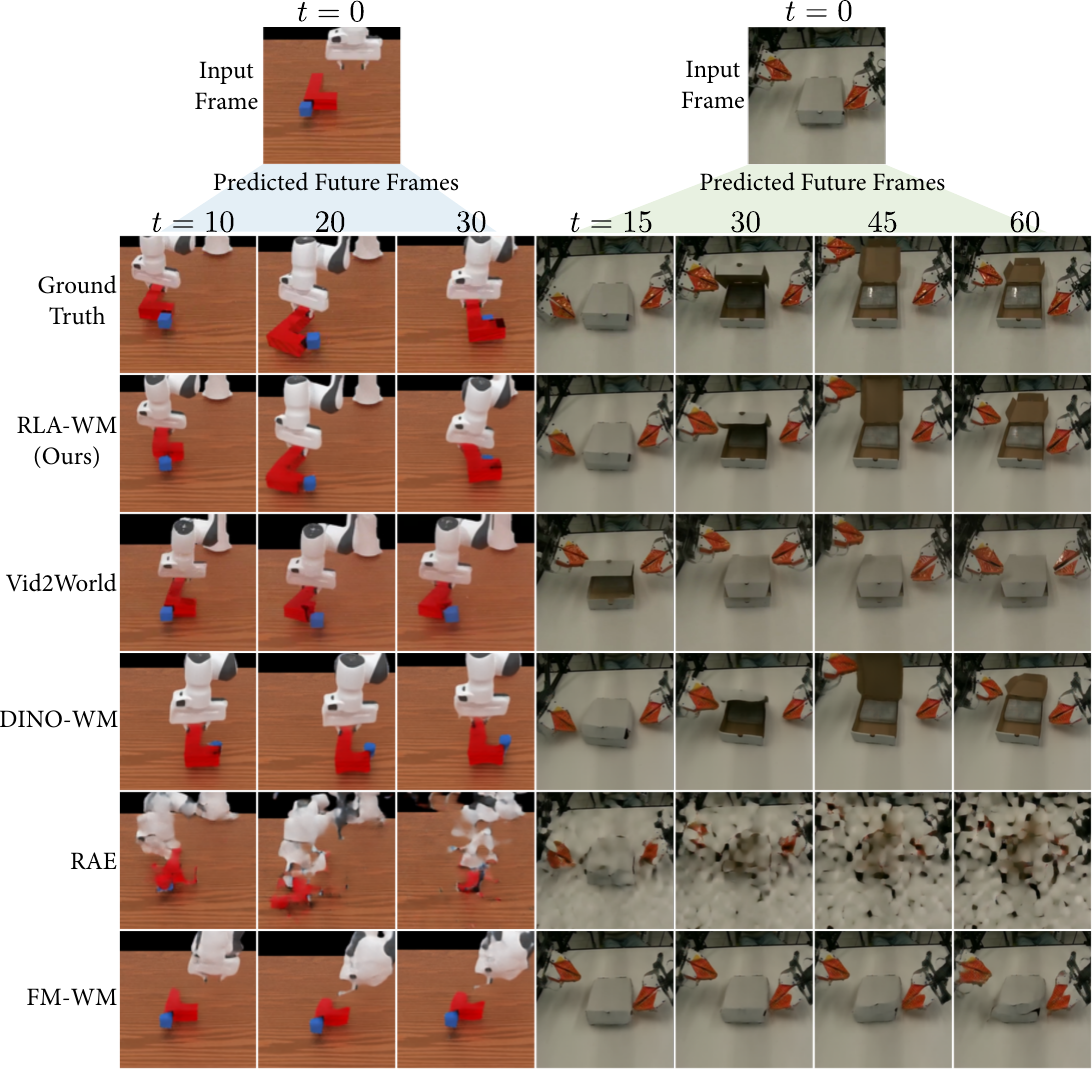}}
    \caption{\textbf{Additional Qualitative Comparison for RLA-WM.}}
    \label{fig:wm-result-4}
\end{figure}

\begin{figure}[h]
    \centering
     \makebox[\linewidth][l]{%
        \hspace*{-1.5em}
    \includegraphics[width=\linewidth]{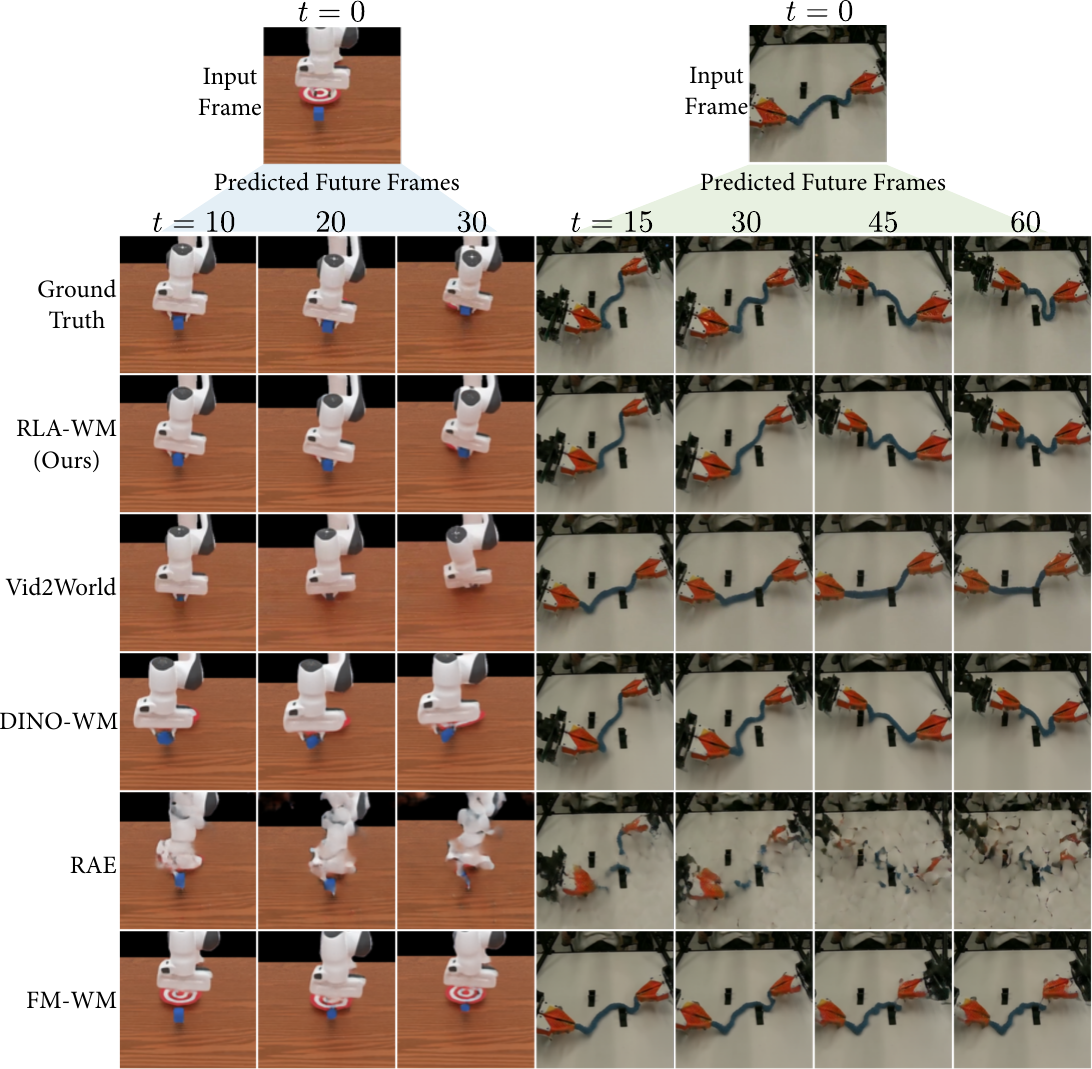}}
    \caption{\textbf{Additional Qualitative Comparison for RLA-WM.}}
    \label{fig:wm-result-5}
\end{figure}



\begin{thebibliography}{10}

\bibitem{dong2026learning}
Jiahua Dong, Qi~Lyu, Baichen Liu, Xudong Wang, Wenqi Liang, Duzhen Zhang, Jiahang Tu, Hongliu Li, Hanbin Zhao, Henghui Ding, et~al.
\newblock Learning to model the world: A survey of world models in artificial intelligence.
\newblock 2026.

\bibitem{yang2024learning}
Mengjiao Yang, Yilun Du, Kamyar Ghasemipour, Jonathan Tompson, Dale Schuurmans, and Pieter Abbeel.
\newblock Learning interactive real-world simulators.
\newblock In {\em International Conference on Learning Representations (ICLR)}, 2024.

\bibitem{wu2025rlvr}
Jialong Wu, Shaofeng Yin, Ningya Feng, and Mingsheng Long.
\newblock Rlvr-world: Training world models with reinforcement learning.
\newblock {\em arXiv preprint arXiv:2505.13934}, 2025.

\bibitem{wu2025geometry}
Haoyu Wu, Diankun Wu, Tianyu He, Junliang Guo, Yang Ye, Yueqi Duan, and Jiang Bian.
\newblock Geometry forcing: Marrying video diffusion and 3d representation for consistent world modeling.
\newblock {\em arXiv preprint arXiv:2507.07982}, 2025.

\bibitem{wang2025precise}
Yuang Wang, Chao Wen, Haoyu Guo, Sida Peng, Minghan Qin, Hujun Bao, Xiaowei Zhou, and Ruizhen Hu.
\newblock Precise action-to-video generation through visual action prompts.
\newblock In {\em Proceedings of the IEEE/CVF International Conference on Computer Vision}, pages 12713--12724, 2025.

\bibitem{huang2025vid2world}
Siqiao Huang, Jialong Wu, Qixing Zhou, Shangchen Miao, and Mingsheng Long.
\newblock Vid2world: Crafting video diffusion models to interactive world models.
\newblock {\em arXiv preprint arXiv:2505.14357}, 2025.

\bibitem{cen2025worldvla}
Jun Cen, Chaohui Yu, Hangjie Yuan, Yuming Jiang, Siteng Huang, Jiayan Guo, Xin Li, Yibing Song, Hao Luo, Fan Wang, et~al.
\newblock Worldvla: Towards autoregressive action world model.
\newblock {\em arXiv preprint arXiv:2506.21539}, 2025.

\bibitem{han2025video}
Hui Han, Siyuan Li, Jiaqi Chen, Yiwen Yuan, Yuling Wu, Yufan Deng, Chak~Tou Leong, Hanwen Du, Junchen Fu, Youhua Li, et~al.
\newblock Video-bench: Human-aligned video generation benchmark.
\newblock In {\em Proceedings of the Computer Vision and Pattern Recognition Conference}, pages 18858--18868, 2025.

\bibitem{zheng2025open}
Zangwei Zheng, Xiangyu Peng, Yuxuan Lou, Chenhui Shen, Tom Young, Xinying Guo, Binluo Wang, Hang Xu, Hongxin Liu, Mingyan Jiang, et~al.
\newblock Open-sora 2.0: Training a commercial-level video generation model in \$200 k.
\newblock {\em arXiv preprint arXiv:2503.09642}, 2025.

\bibitem{wang2026interactive}
Yixuan Wang, Rhythm Syed, Fangyu Wu, Mengchao Zhang, Aykut Onol, Jose Barreiros, Hooshang Nayyeri, Tony Dear, Huan Zhang, and Yunzhu Li.
\newblock Interactive world simulator for robot policy training and evaluation.
\newblock {\em arXiv preprint arXiv:2603.08546}, 2026.

\bibitem{team2025gigabrain}
GigaBrain Team, Angen Ye, Boyuan Wang, Chaojun Ni, Guan Huang, Guosheng Zhao, Haoyun Li, Jie Li, Jiagang Zhu, Lv~Feng, et~al.
\newblock Gigabrain-0: A world model-powered vision-language-action model.
\newblock {\em arXiv preprint arXiv:2510.19430}, 2025.

\bibitem{mendonca2023structured}
Russell Mendonca, Shikhar Bahl, and Deepak Pathak.
\newblock Structured world models from human videos.
\newblock In {\em Robotics: Science and Systems (RSS)}, 2023.

\bibitem{lu2025gwm}
Guanxing Lu, Baoxiong Jia, Puhao Li, Yixin Chen, Ziwei Wang, Yansong Tang, and Siyuan Huang.
\newblock Gwm: Towards scalable gaussian world models for robotic manipulation.
\newblock In {\em Proceedings of the IEEE/CVF International Conference on Computer Vision}, pages 9263--9274, 2025.

\bibitem{du2024video}
Yilun Du, Mengjiao Yang, Pete Florence, Fei Xia, Ayzaan Wahid, Brian Ichter, Pierre Sermanet, Tianhe Yu, Pieter Abbeel, Joshua~B Tenenbaum, et~al.
\newblock Video language planning.
\newblock In {\em International Conference on Learning Representations (ICLR)}, 2024.

\bibitem{chen2025planning}
Delong Chen, Theo Moutakanni, Willy Chung, Yejin Bang, Ziwei Ji, Allen Bolourchi, and Pascale Fung.
\newblock Planning with reasoning using vision language world model.
\newblock {\em arXiv preprint arXiv:2509.02722}, 2025.

\bibitem{zhou2025dino}
Gaoyue Zhou, Hengkai Pan, Yann LeCun, and Lerrel Pinto.
\newblock Dino-wm: World models on pre-trained visual features enable zero-shot planning.
\newblock In {\em International Conference on Machine Learning (ICML)}, 2025.

\bibitem{assran2025v}
Mido Assran, Adrien Bardes, David Fan, Quentin Garrido, Russell Howes, Matthew Muckley, Ammar Rizvi, Claire Roberts, Koustuv Sinha, Artem Zholus, et~al.
\newblock V-jepa 2: Self-supervised video models enable understanding, prediction and planning.
\newblock {\em arXiv preprint arXiv:2506.09985}, 2025.

\bibitem{yamins2016using}
Daniel~LK Yamins and James~J DiCarlo.
\newblock Using goal-driven deep learning models to understand sensory cortex.
\newblock {\em Nature neuroscience}, 19(3):356--365, 2016.

\bibitem{battaglia2013simulation}
Peter~W Battaglia, Jessica~B Hamrick, and Joshua~B Tenenbaum.
\newblock Simulation as an engine of physical scene understanding.
\newblock {\em Proceedings of the national academy of sciences}, 110(45):18327--18332, 2013.

\bibitem{chun2025sparse}
Junha Chun, Youngjoon Jeong, and Taesup Kim.
\newblock Sparse imagination for efficient visual world model planning.
\newblock {\em arXiv preprint arXiv:2506.01392}, 2025.

\bibitem{baldassarre2025back}
Federico Baldassarre, Marc Szafraniec, Basile Terver, Vasil Khalidov, Francisco Massa, Yann LeCun, Patrick Labatut, Maximilian Seitzer, and Piotr Bojanowski.
\newblock Back to the features: Dino as a foundation for video world models.
\newblock {\em arXiv preprint arXiv:2507.19468}, 2025.

\bibitem{zheng2025rethinking}
Zhenxin Zheng and Zhenjie Zheng.
\newblock Rethinking diffusion model in high dimension.
\newblock {\em arXiv preprint arXiv:2503.08643}, 2025.

\bibitem{zheng2025diffusion}
Boyang Zheng, Nanye Ma, Shengbang Tong, and Saining Xie.
\newblock Diffusion transformers with representation autoencoders.
\newblock {\em arXiv preprint arXiv:2510.11690}, 2025.

\bibitem{gao2025adaworld}
Shenyuan Gao, Siyuan Zhou, Yilun Du, Jun Zhang, and Chuang Gan.
\newblock Adaworld: Learning adaptable world models with latent actions.
\newblock In {\em International Conference on Machine Learning (ICML)}, 2025.

\bibitem{bi2025motus}
Hongzhe Bi, Hengkai Tan, Shenghao Xie, Zeyuan Wang, Shuhe Huang, Haitian Liu, Ruowen Zhao, Yao Feng, Chendong Xiang, Yinze Rong, et~al.
\newblock Motus: A unified latent action world model.
\newblock {\em arXiv preprint arXiv:2512.13030}, 2025.

\bibitem{liu2023libero}
Bo~Liu, Yifeng Zhu, Chongkai Gao, Yihao Feng, Qiang Liu, Yuke Zhu, and Peter Stone.
\newblock {LIBERO}: Benchmarking knowledge transfer for lifelong robot learning.
\newblock In {\em Advances in Neural Information Processing Systems}, volume~36, 2023.

\bibitem{zhen20243d}
Haoyu Zhen, Xiaowen Qiu, Peihao Chen, Jincheng Yang, Xin Yan, Yilun Du, Yining Hong, and Chuang Gan.
\newblock 3d-vla: A 3d vision-language-action generative world model.
\newblock In {\em International Conference on Machine Learning (ICML)}, 2024.

\bibitem{yu2025manigaussian++}
Tengbo Yu, Guanxing Lu, Zaijia Yang, Haoyuan Deng, Season~Si Chen, Jiwen Lu, Wenbo Ding, Guoqiang Hu, Yansong Tang, and Ziwei Wang.
\newblock Manigaussian++: General robotic bimanual manipulation with hierarchical gaussian world model.
\newblock In {\em 2025 IEEE/RSJ International Conference on Intelligent Robots and Systems (IROS)}, pages 12232--12239. IEEE, 2025.

\bibitem{huang2026pointworld}
Wenlong Huang, Yu-Wei Chao, Arsalan Mousavian, Ming-Yu Liu, Dieter Fox, Kaichun Mo, and Li~Fei-Fei.
\newblock Pointworld: Scaling 3d world models for in-the-wild robotic manipulation.
\newblock {\em arXiv preprint arXiv:2601.03782}, 2026.

\bibitem{huang2025particleformer}
Suning Huang, Qianzhong Chen, Xiaohan Zhang, Jiankai Sun, and Mac Schwager.
\newblock Particleformer: A 3d point cloud world model for multi-object, multi-material robotic manipulation.
\newblock {\em arXiv preprint arXiv:2506.23126}, 2025.

\bibitem{chai2025gaf}
Ying Chai, Litao Deng, Ruizhi Shao, Jiajun Zhang, Kangchen Lv, Liangjun Xing, Xiang Li, Hongwen Zhang, and Yebin Liu.
\newblock Gaf: Gaussian action field as a 4d representation for dynamic world modeling in robotic manipulation.
\newblock {\em arXiv preprint arXiv:2506.14135}, 2025.

\bibitem{li2025robotic}
Chenhao Li, Andreas Krause, and Marco Hutter.
\newblock Robotic world model: A neural network simulator for robust policy optimization in robotics.
\newblock {\em arXiv preprint arXiv:2501.10100}, 2025.

\bibitem{jyothir2023gradient}
SV~Jyothir, Siddhartha Jalagam, Yann LeCun, and Vlad Sobal.
\newblock Gradient-based planning with world models.
\newblock {\em arXiv preprint arXiv:2312.17227}, pages 703--708, 2023.

\bibitem{hansen2024td}
Nicklas Hansen, Hao Su, and Xiaolong Wang.
\newblock Td-mpc2: Scalable, robust world models for continuous control.
\newblock In {\em International Conference on Learning Representations (ICLR)}, 2024.

\bibitem{hafner2023mastering}
Danijar Hafner, Jurgis Pasukonis, Jimmy Ba, and Timothy Lillicrap.
\newblock Mastering diverse domains through world models.
\newblock {\em arXiv preprint arXiv:2301.04104}, 2023.

\bibitem{hafner2025training}
Danijar Hafner, Wilson Yan, and Timothy Lillicrap.
\newblock Training agents inside of scalable world models.
\newblock {\em arXiv preprint arXiv:2509.24527}, 2025.

\bibitem{georgiev2025pwm}
Ignat Georgiev, Varun Giridhar, Nicklas Hansen, and Animesh Garg.
\newblock Pwm: Policy learning with multi-task world models.
\newblock In {\em International Conference on Learning Representations (ICLR)}, 2025.

\bibitem{mur2026v}
Lorenzo Mur-Labadia, Matthew Muckley, Amir Bar, Mido Assran, Koustuv Sinha, Mike Rabbat, Yann LeCun, Nicolas Ballas, and Adrien Bardes.
\newblock V-jepa 2.1: Unlocking dense features in video self-supervised learning.
\newblock {\em arXiv preprint arXiv:2603.14482}, 2026.

\bibitem{oquab2023dinov2}
Maxime Oquab, Timoth{\'e}e Darcet, Th{\'e}o Moutakanni, Huy Vo, Marc Szafraniec, Vasil Khalidov, Pierre Fernandez, Daniel Haziza, Francisco Massa, Alaaeldin El-Nouby, et~al.
\newblock Dinov2: Learning robust visual features without supervision.
\newblock {\em arXiv preprint arXiv:2304.07193}, 2023.

\bibitem{simeoni2025dinov3}
Oriane Sim{\'e}oni, Huy~V Vo, Maximilian Seitzer, Federico Baldassarre, Maxime Oquab, Cijo Jose, Vasil Khalidov, Marc Szafraniec, Seungeun Yi, Micha{\"e}l Ramamonjisoa, et~al.
\newblock Dinov3.
\newblock {\em arXiv preprint arXiv:2508.10104}, 2025.

\bibitem{zhang2025latent}
Chuheng Zhang, Tim Pearce, Pushi Zhang, Kaixin Wang, Xiaoyu Chen, Wei Shen, Li~Zhao, and Jiang Bian.
\newblock What do latent action models actually learn?
\newblock {\em arXiv preprint arXiv:2506.15691}, 2025.

\bibitem{zheng2025flare}
Ruijie Zheng, Jing Wang, Scott Reed, Johan Bjorck, Yu~Fang, Fengyuan Hu, Joel Jang, Kaushil Kundalia, Zongyu Lin, Loic Magne, et~al.
\newblock Flare: Robot learning with implicit world modeling.
\newblock {\em arXiv preprint arXiv:2505.15659}, 2025.

\bibitem{ye2025latent}
Seonghyeon Ye, Joel Jang, Byeongguk Jeon, Sejune Joo, Jianwei Yang, Baolin Peng, Ajay Mandlekar, Reuben Tan, Yu-Wei Chao, Bill~Yuchen Lin, et~al.
\newblock Latent action pretraining from videos.
\newblock In {\em International Conference on Learning Representations (ICLR)}, 2025.

\bibitem{tharwat2025latent}
Bahey Tharwat, Yara Nasser, Ali Abouzeid, and Ian Reid.
\newblock Latent action pretraining through world modeling.
\newblock {\em arXiv preprint arXiv:2509.18428}, 2025.

\bibitem{bu2025univla}
Qingwen Bu, Yanting Yang, Jisong Cai, Shenyuan Gao, Guanghui Ren, Maoqing Yao, Ping Luo, and Hongyang Li.
\newblock Univla: Learning to act anywhere with task-centric latent actions.
\newblock {\em arXiv preprint arXiv:2505.06111}, 2025.

\bibitem{kerssies2026frame}
Tommie Kerssies, Gabriele Berton, Ju~He, Qihang Yu, Wufei Ma, Daan de~Geus, Gijs Dubbelman, and Liang-Chieh Chen.
\newblock A frame is worth one token: Efficient generative world modeling with delta tokens.
\newblock In {\em Proceedings of the IEEE/CVF Conference on Computer Vision and Pattern Recognition}, pages 27978--27988, 2026.

\bibitem{esser2024scaling}
Patrick Esser, Sumith Kulal, Andreas Blattmann, Rahim Entezari, Jonas M{\"u}ller, Harry Saini, Yam Levi, Dominik Lorenz, Axel Sauer, Frederic Boesel, et~al.
\newblock Scaling rectified flow transformers for high-resolution image synthesis.
\newblock In {\em Forty-first international conference on machine learning}, 2024.

\bibitem{shi2023diffusion}
Yuyang Shi, Valentin De~Bortoli, Andrew Campbell, and Arnaud Doucet.
\newblock Diffusion schr{\"o}dinger bridge matching.
\newblock {\em Advances in neural information processing systems}, 36:62183--62223, 2023.

\bibitem{de2024schrodinger}
Valentin De~Bortoli, Iryna Korshunova, Andriy Mnih, and Arnaud Doucet.
\newblock Schrodinger bridge flow for unpaired data translation.
\newblock {\em Advances in Neural Information Processing Systems}, 37:103384--103441, 2024.

\bibitem{chen2022crom}
Peter~Yichen Chen, Jinxu Xiang, Dong~Heon Cho, Yue Chang, GA~Pershing, Henrique~Teles Maia, Maurizio~M Chiaramonte, Kevin Carlberg, and Eitan Grinspun.
\newblock Crom: Continuous reduced-order modeling of pdes using implicit neural representations.
\newblock {\em arXiv preprint arXiv:2206.02607}, 2022.

\bibitem{mu2maniskill}
Tongzhou Mu, Zhan Ling, Fanbo Xiang, Derek~Cathera Yang, Xuanlin Li, Stone Tao, Zhiao Huang, Zhiwei Jia, and Hao Su.
\newblock Maniskill: Generalizable manipulation skill benchmark with large-scale demonstrations.
\newblock In {\em Thirty-fifth Conference on Neural Information Processing Systems Datasets and Benchmarks Track (Round 2)}.

\bibitem{lipmanflow}
Yaron Lipman, Ricky~TQ Chen, Heli Ben-Hamu, Maximilian Nickel, and Matthew Le.
\newblock Flow matching for generative modeling.
\newblock In {\em The Eleventh International Conference on Learning Representations}.

\bibitem{zhang2018lpips}
Richard Zhang, Phillip Isola, Alexei~A Efros, Eli Shechtman, and Oliver Wang.
\newblock The unreasonable effectiveness of deep features as a perceptual metric.
\newblock In {\em Proceedings of the IEEE conference on computer vision and pattern recognition}, pages 586--595, 2018.

\bibitem{wang2004image}
Zhou Wang, Alan~C Bovik, Hamid~R Sheikh, and Eero~P Simoncelli.
\newblock Image quality assessment: from error visibility to structural similarity.
\newblock {\em IEEE transactions on image processing}, 13(4):600--612, 2004.

\bibitem{xing2024dynamicrafter}
Jinbo Xing, Menghan Xia, Yong Zhang, Haoxin Chen, Wangbo Yu, Hanyuan Liu, Gongye Liu, Xintao Wang, Ying Shan, and Tien-Tsin Wong.
\newblock Dynamicrafter: Animating open-domain images with video diffusion priors.
\newblock In {\em European Conference on Computer Vision}, pages 399--417. Springer, 2024.

\bibitem{ronneberger2015u}
Olaf Ronneberger, Philipp Fischer, and Thomas Brox.
\newblock U-net: Convolutional networks for biomedical image segmentation.
\newblock In {\em International Conference on Medical image computing and computer-assisted intervention}, pages 234--241. Springer, 2015.

\bibitem{yuan2026fastwam}
Tianyuan Yuan, Zibin Dong, Yicheng Liu, and Hang Zhao.
\newblock Fast-wam: Do world action models need test-time future imagination?
\newblock {\em arXiv preprint arXiv:2603.16666}, 2026.

\bibitem{he2016deep}
Kaiming He, Xiangyu Zhang, Shaoqing Ren, and Jian Sun.
\newblock Deep residual learning for image recognition.
\newblock In {\em Proceedings of the IEEE conference on computer vision and pattern recognition}, pages 770--778, 2016.

\bibitem{van2017neural}
Aaron Van Den~Oord, Oriol Vinyals, et~al.
\newblock Neural discrete representation learning.
\newblock {\em Advances in neural information processing systems}, 30, 2017.

\bibitem{wang2026vla}
Yihao Wang, Pengxiang Ding, Lingxiao Li, Can Cui, Zirui Ge, Xinyang Tong, Wenxuan Song, Han Zhao, Wei Zhao, Pengxu Hou, et~al.
\newblock Vla-adapter: An effective paradigm for tiny-scale vision-language-action model.
\newblock In {\em Proceedings of the AAAI conference on artificial intelligence}, volume~40, pages 18638--18646, 2026.

\bibitem{ha2018world}
David Ha and J{\"u}rgen Schmidhuber.
\newblock World models.
\newblock {\em arXiv preprint arXiv:1803.10122}, 2(3):440, 2018.

\bibitem{berg2025semantic}
Jacob Berg, Chuning Zhu, Yanda Bao, Ishan Durugkar, and Abhishek Gupta.
\newblock Semantic world models.
\newblock {\em arXiv preprint arXiv:2510.19818}, 2025.

\bibitem{hansen2022temporal}
Nicklas Hansen, Xiaolong Wang, and Hao Su.
\newblock Temporal difference learning for model predictive control.
\newblock {\em arXiv preprint arXiv:2203.04955}, 2022.

\bibitem{haarnoja2018soft}
Tuomas Haarnoja, Aurick Zhou, Pieter Abbeel, and Sergey Levine.
\newblock Soft actor-critic: Off-policy maximum entropy deep reinforcement learning with a stochastic actor.
\newblock In {\em International conference on machine learning}, pages 1861--1870. PMLR, 2018.

\bibitem{lireinforcement}
Qiyang Li, Zhiyuan Zhou, and Sergey Levine.
\newblock Reinforcement learning with action chunking.
\newblock In {\em The Thirty-ninth Annual Conference on Neural Information Processing Systems}.

\bibitem{litop}
Ge~Li, Dong Tian, Hongyi Zhou, Xinkai Jiang, Rudolf Lioutikov, and Gerhard Neumann.
\newblock Top-erl: Transformer-based off-policy episodic reinforcement learning.

\bibitem{schulman2017proximal}
John Schulman, Filip Wolski, Prafulla Dhariwal, Alec Radford, and Oleg Klimov.
\newblock Proximal policy optimization algorithms.
\newblock {\em arXiv preprint arXiv:1707.06347}, 2017.

\bibitem{levine2020offline}
Sergey Levine, Aviral Kumar, George Tucker, and Justin Fu.
\newblock Offline reinforcement learning: Tutorial, review, and perspectives on open problems.
\newblock {\em arXiv preprint arXiv:2005.01643}, 2020.

\bibitem{foster2024behavior}
Dylan~J Foster, Adam Block, and Dipendra Misra.
\newblock Is behavior cloning all you need? understanding horizon in imitation learning.
\newblock {\em Advances in Neural Information Processing Systems}, 37:120602--120666, 2024.

\bibitem{henderson2018deep}
Peter Henderson, Riashat Islam, Philip Bachman, Joelle Pineau, Doina Precup, and David Meger.
\newblock Deep reinforcement learning that matters.
\newblock In {\em Proceedings of the AAAI conference on artificial intelligence}, volume~32, 2018.

\bibitem{mania2018simple}
Horia Mania, Aurelia Guy, and Benjamin Recht.
\newblock Simple random search provides a competitive approach to reinforcement learning.
\newblock {\em arXiv preprint arXiv:1803.07055}, 2018.

\bibitem{loshchilovdecoupled}
Ilya Loshchilov and Frank Hutter.
\newblock Decoupled weight decay regularization.
\newblock In {\em International Conference on Learning Representations}.

\bibitem{xiang2025structured}
Jianfeng Xiang, Zelong Lv, Sicheng Xu, Yu~Deng, Ruicheng Wang, Bowen Zhang, Dong Chen, Xin Tong, and Jiaolong Yang.
\newblock Structured 3d latents for scalable and versatile 3d generation.
\newblock In {\em Proceedings of the IEEE/CVF conference on computer vision and pattern recognition}, pages 21469--21480, 2025.

\bibitem{chi2025diffusion}
Cheng Chi, Zhenjia Xu, Siyuan Feng, Eric Cousineau, Yilun Du, Benjamin Burchfiel, Russ Tedrake, and Shuran Song.
\newblock Diffusion policy: Visuomotor policy learning via action diffusion.
\newblock {\em The International Journal of Robotics Research}, 44(10-11):1684--1704, 2025.

\bibitem{schulman2015high}
John Schulman, Philipp Moritz, Sergey Levine, Michael Jordan, and Pieter Abbeel.
\newblock High-dimensional continuous control using generalized advantage estimation.
\newblock {\em arXiv preprint arXiv:1506.02438}, 2015.

\bibitem{spearman1961proof}
Charles Spearman.
\newblock The proof and measurement of association between two things.
\newblock 1961.

\bibitem{zhang2024cameras}
Jason~Y. Zhang, Amy Lin, Moneish Kumar, Tzu-Hsuan Yang, Deva Ramanan, and Shubham Tulsiani.
\newblock Cameras as rays: Pose estimation via ray diffusion.
\newblock In {\em International Conference on Learning Representations}, 2024.

\bibitem{lillicrap2015continuous}
Timothy~P Lillicrap, Jonathan~J Hunt, Alexander Pritzel, Nicolas Heess, Tom Erez, Yuval Tassa, David Silver, and Daan Wierstra.
\newblock Continuous control with deep reinforcement learning.
\newblock {\em arXiv preprint arXiv:1509.02971}, 2015.

\end{thebibliography}
\end{document}